\documentclass{article}

\PassOptionsToPackage{numbers, compress}{natbib}

\usepackage{arydshln}

\usepackage[main, final]{neurips_2026}

\usepackage[utf8]{inputenc} 
\usepackage[T1]{fontenc}    
\usepackage{hyperref}       
\usepackage{url}            
\usepackage{booktabs}       
\usepackage{amsfonts}       
\usepackage{nicefrac}       
\usepackage{microtype}      
\usepackage{xcolor}         
\usepackage{amsmath}
\usepackage{graphicx}
\usepackage{booktabs}
\usepackage{multirow}
\usepackage[table]{xcolor}
\usepackage{booktabs}
\usepackage{tabularx}
\usepackage{array}

\usepackage{booktabs}
\usepackage{tabularx}
\usepackage{xcolor}
\usepackage{array}

\usepackage{wrapfig}
\usepackage{needspace}

\usepackage{threeparttable}

\definecolor{increasecolor}{RGB}{34, 113, 181}
\definecolor{decreasecolor}{RGB}{217, 95, 14}
\newcommand{\poschg}[1]{\textcolor{increasecolor}{\textbf{#1}}}
\newcommand{\negchg}[1]{\textcolor{decreasecolor}{\textbf{#1}}}

\usepackage[most]{tcolorbox}
\usepackage{listings}
\usepackage{xcolor}
\usepackage{caption}
\tcbuselibrary{breakable}

\definecolor{m2aGreen}{HTML}{2E7D32}
\definecolor{m2aBlue}{HTML}{1565C0}
\definecolor{m2aOrange}{HTML}{EF6C00}
\definecolor{m2aRed}{HTML}{C62828}
\definecolor{m2aGray}{HTML}{F5F5F5}
\definecolor{m2aDarkGray}{HTML}{616161}

\lstdefinestyle{trajstyle}{
  basicstyle=\ttfamily\scriptsize,
  breaklines=true,
  columns=fullflexible,
  keepspaces=true,
  showstringspaces=false,
  frame=none,
  xleftmargin=0pt,
  xrightmargin=0pt,
  aboveskip=2pt,
  belowskip=2pt
}

\newcommand{\trajtag}[2]{
  \tcbox[
    on line,
    colback=#1!10,
    colframe=#1!65!black,
    boxrule=0.35pt,
    arc=2pt,
    left=2pt,
    right=2pt,
    top=1pt,
    bottom=1pt
  ]{\scriptsize\bfseries #2}
}

\usepackage{xcolor}
\usepackage{listings}

\lstdefinestyle{trajcode}{
    basicstyle=\ttfamily\small,
    breaklines=true,
    columns=fullflexible,
    keepspaces=true,
    frame=single,
    framerule=0.3pt,
    rulecolor=\color{black!30},
    backgroundcolor=\color{gray!5},
    showstringspaces=false,
    tabsize=2
}

\title{M2A: Synergizing Mathematical and Agentic Reasoning in Large Language Models}

\author{%
  Junjian Wang$^{1,2}$%
  \quad
  Xin Zhou$^{3}$%
  \thanks{Corresponding author.}%
  \quad
  Qiran Xu$^{3}$%
  \quad
  Kun Zhan$^{3}$%
  \\[0.5em]
  {\normalfont\small
  $^{1}$Foundation Model Research Center, Institute of Automation, Chinese Academy of Sciences
  }\\[0.2em]
  {\normalfont\small
  $^{2}$School of Artificial Intelligence, University of Chinese Academy of Sciences
  }\\[0.2em]
  {\normalfont\small
  $^{3}$Li Auto Inc.
  }\\[0.4em]
  {\normalfont\small
  \texttt{wangjunjian2025@ia.ac.cn}
  \quad
  \texttt{\{zhouxin12,xuqiran,zhankun\}@lixiang.com}
  }%
}

\begin{document}

\maketitle
\begingroup
\renewcommand{\thefootnote}{}
\footnotetext{Code is available at
\url{https://github.com/Laplucky/M2A}.}
\endgroup

\begin{abstract}

While reasoning has become a central capability of large language models (LLMs), the reasoning patterns required for different scenarios are often misaligned.
Mathematical reasoning typically relies on intrinsic logic to solve closed-world problems in a single response, whereas agentic reasoning requires not only internal reasoning but also multi-turn interaction with external environments, interleaving thought and action.
This misalignment prevents mathematical and agentic reasoning from effectively benefiting from each other, often yielding unstable reasoning behavior and only limited performance gains under multi-task learning.
In this paper, we propose M2A, a novel paradigm that synergizes mathematical and agentic reasoning via model merging. To avoid overfitting to superficial reasoning patterns under joint training, M2A operates directly in parameter space: it identifies the feature subspace critical for agent behavior, and merges the mathematical reasoning task vector only along its null space, thereby injecting reasoning capability along directions that do not perturb agent behavior.
Unlike SFT or RL, M2A requires no additional gradient-update and exposes the merging coefficient as a simple knob for controlling reasoning length.
Experiments in a challenging real-world coding agent setting show that our method effectively extends agentic reasoning depth and delivers substantial performance improvements. 
Applied to a fine-tuned Qwen3-8B, M2A improves its SWE-Bench Verified resolved rate from 44.0\% to 51.2\% without retraining the model. Further experiments demonstrate that M2A generalizes
across model families, agent scaffolds, and task domains.

\end{abstract}

\section{Introduction}
\label{sec:introduction}

Reasoning serves as the core of human intelligence, driving critical processes from problem-solving to strategic planning. In the realm of Large Language Models (LLMs), chain-of-thought prompting ~\citep{wei2022chain,kojima2022large} shows that eliciting reasoning can unlock substantial gains on complex tasks, and recent  large reasoning models~\citep{jaech2024openai,guo2025deepseek, comanici2025gemini} further advanced \textit{mathematical reasoning} capability through reinforcement learning with verifiable rewards (RLVR), achieving strong performance in mathematics, competitive programming, and broader STEM domains. 
More recently, the frontier of reasoning has shifted beyond closed-world problem solving toward \emph{agentic reasoning}, where models must reason while interacting with external environments over multiple turns \cite{wei2026agentic}. Prior work such as ReAct~\citep{yao2022react} highlights that tightly coupling reasoning with action is essential for complex interactive tasks, making the improvement of agentic reasoning increasingly important for LLM-based agents.

Although reasoning is crucial across many tasks, the behavior patterns required by different reasoning scenarios are often misaligned, particularly between mathematical reasoning and agentic reasoning. 
Mathematical reasoning is primarily a process of \textit{internal thinking}: the model works in a relatively closed-world setting and aims to produce a complete solution in a single response \cite{pang2025reasoning, balunovic2025matharena, huan2025does}. 
In contrast, agentic reasoning is better characterized as a process of \textit{coordinating internal thinking with external action}: the model must keep interacting with an open-world environment over multiple turns and decide when to think and when to act \cite{yao2022react, shinn2023reflexion, wei2026agentic}. 
Such contrasts across task domains, interaction formats, and reasoning styles prevent multi-task learning from improving both capabilities together, and can even let mathematical reasoning patterns override agentic behaviors.
Recent studies \cite{huan2025does,li2026reasoning,wang2026mix} and our experiments (Table \ref{tab:main_results_full}) show a consistent pattern: multi-task SFT \cite{huan2025does} or RL \cite{wang2026mix} boosts mathematical reasoning but barely helps on agent tasks.

In this paper, we present M2A, a novel paradigm that synergizes mathematical and agentic reasoning in a training-free manner. 
To avoid fitting mismatched data distribution during multi-task training, \textbf{M2A couples the two capabilities directly in parameter space through model merging}. While this mitigates the training-stage mismatch, finer-grained conflicts still remain in parameter space.
Naively merging~\citep{ilharcoediting, yadav2023ties, goddard2024arcee} a reasoning model into an agent model can overwrite the agent's interaction behavior: the merged model reasons for substantially longer but interacts with the environment far less frequently, reducing external actions and ultimately degrading agent performance. 
This observation motivates us to reformulate our goal as a \textbf{behavior-preserving capability integration problem}—how to combine mathematical and agentic reasoning without disrupting the agent's interaction pattern. 
Taking inspiration from \cite{wang2021training, fang2024alphaedit, huang2026rain}, M2A identifies the parameter subspace critical for agent behavior and applies null-space projection to the mathematical-reasoning update before merging it into the agent model, thereby integrating the two capabilities while minimally perturbing agentic behavior. 
To further mitigate capability conflicts at a finer granularity, \textbf{we introduce an adaptive layer-wise merging coefficient mechanism}, which additionally serves as a natural knob for adjusting reasoning length on top of existing models. 
Compared with costly SFT~\citep{wang2025swe, tao2026swe} or RL~\citep{sun2026swe, lian2026swe}, M2A requires no further training to the existing model, and offers a simple yet effective mechanism for controlling reasoning length.

We evaluate M2A and baselines on SWE-Bench Verified~\citep{jimenez2023swe}, a challenging coding-agent benchmark that requires resolving real-world GitHub issues. 
Our experiments show that multi-task SFT, while able to transfer superficial reasoning traces, fails to translate them into agent capability and can actively hurt agent performance. In contrast, M2A improves the Qwen3-8B baseline from 44.0\% to 51.2\% on SWE-Bench Verified, while growing the average per-step reasoning length from 253.3 to 327.4 tokens. 
More importantly, M2A makes this transfer controllable: we observe that the average reasoning length grows monotonically and nearly linear with the merging coefficient, turning the coefficient into a predictable knob for modulating agent reasoning behavior. 
The ablation study demonstrates the efficacy of each module within the M2A framework. Our analysis further reveals that M2A not only enhances per-step reasoning depth but also shifts the agent’s trajectory-level behavior toward more frequent external actions.
Additional experiments on Llama-3.1-8B, Mini-SWE-Agent \cite{yang2024swe},
R2E \cite{jain2025r2e}, and GAIA \cite{mialon2024gaia} further demonstrate the generalizability
of M2A across model families, agent scaffolds, and
task domains.

Our contributions are as follows: 
(1) We propose M2A, a novel paradigm that synergizes mathematical reasoning and agentic reasoning in LLMs in a training-free manner. M2A preserves the agent's external action behavior while injecting stronger internal reasoning through behavior-preserving null-space model merging. 
(2) On SWE-Bench Verified, M2A improves a fine-tuned Qwen3-8B from 44.0\% to 51.2\% resolved rate, consistently outperforming multi-task SFT and representative model-merging baselines. 
(3) We conduct systematic ablations and analysis to demonstrate the effectiveness of M2A, and show that merge strength serves as an effective control knob over reasoning length, enabling explicit control of the reasoning behavior.

\section{Related Work}
\subsection{Generalization of LLM Reasoning}
LLM reasoning has become a central mechanism for improving complex problem solving~\citep{wei2022chain,jaech2024openai,guo2025deepseek}. 
However, reasoning gains do not always generalize beyond the training domain~\citep{bai2025how,huan2025does}. 
Although mathematical reasoning is widely viewed as a general source of reasoning capability, math-specialized models often bring limited benefits to broader tasks such as scientific QA, coding, and agent planning~\citep{huan2025does,bai2025how}.
Recent work therefore explores broader reasoning data and multi-domain RL objectives~\citep{su2025crossing,ma2025general,pang2025reasoning,ramesh2026multitask,wang2026mix}, but the resulting generalization remains sensitive to the target behavior.
This limitation is especially pronounced for agentic reasoning. 
Prior studies show that math reasoning does not consistently improve broader LLM capabilities~\citep{huan2025does}; multi-task RL can improve several reasoning-intensive domains but does not reliably enhance agent performance~\citep{wang2026mix}; and reasoning and tool-use behaviors may interfere under joint optimization~\citep{li2026reasoning}. 
These findings suggest that mathematical and agentic reasoning cannot be reliably combined by
simply mixing their training data. Effective synergy requires strengthening internal thinking while preserving the external action behavior required by agents. This motivates our parameter-level approach, which integrates mathematical reasoning updates through behavior-preserving model merging rather than joint training.

\subsection{Model Merging}
Model merging aims to combine multiple fine-tuned models into a single model without additional training, reducing storage and computational costs \cite{du2024parameter, ma2025led, qi2025less, yang2026model}. Early methods such as Model Soups \cite{wortsman2022model} and Task Arithmetic \cite{ilharcoediting} showed that weight averaging or task-vector composition can be effective when source models are sufficiently aligned. Later approaches, TIES-Merging \cite{yadav2023ties} alleviates parameter interference via pruning, symbol election, and merging, DARE \cite{yu2024language} enhances merged model robustness and base capabilities by randomly dropping and rescaling fine-tuning weights.
Recent studies \cite{yang2026orthogonal, zhang2025unraveling, zheng2025decouple} have explored geometry and subspace-aware formulations to better preserve model structure during composition. However, most model merging methods are primarily designed for relatively aligned capabilities.
RAIN-Merging \cite{huang2026rain} introduces null-space projected merging to preserve the thinking format of reasoning models while enhancing instruction-following ability. But it focuses on preserving the structured thinking format within a single response,
without considering the multi-turn think--act--observe behavior required by agents. Moreover, its instruction-attention-guided merge control
is designed for instruction following and is less applicable to agent tasks.
In contrast, M2A extends null-space projected merging to synergize mathematical and agentic reasoning 
by explicitly calibrating agent-critical behavioral features and adaptively controlling both merge layers and merge strength.



\section{Methodology}


In this section, we first define the problem setting for synergizing mathematical and agentic reasoning, and clarify why traditional multi-task learning and model merging are insufficient. 
We then introduce M2A, a novel paradigm that leverages null-space model merging to integrate mathematical reasoning into an agent model while maintaining its agent behavior pattern.

\subsection{Problem Statement}

We consider a multi-turn ReAct-style agent dataset $\mathcal{D}_{\text{agent}}$ and a single-turn mathematical reasoning dataset $\mathcal{D}_{\text{math}}$. Our goal is to obtain a single model that synergizes both capabilities, achieving strong performance on agent tasks while benefiting from mathematical reasoning. 

\paragraph{Multi-task SFT} A standard approach to combine the two capabilities is joint fine-tuning on $\mathcal{D}=\mathcal{D}_{\text{agent}} \cup \mathcal{D}_{\text{math}}$:
\begin{equation}
\mathcal{L}_{\text{SFT}}(\theta) = -\mathbb{E}_{(x, y) \sim \mathcal{D}} \left[ \sum_{t=1}^{|y|} \log \pi_\theta(y_t \mid x, y_{<t}) \right],
\end{equation}
with the hope that the long chain-of-thought behavior elicited by the math data can generalize to agentic reasoning \cite{wei2022chain, guo2025deepseek}.
However, the two data distributions diverge across task domains, interaction formats, and reasoning styles, causing joint optimization to fit surface-level patterns rather than transferable reasoning capabilities \cite{ma2025general,huan2025does}.
As a result, even if multi-task SFT successfully transfers the surface-level "long reasoning" style, this increased length provides little benefit to—and sometimes even degrades—actual agent performance.

\paragraph{Naive Model Merging} 

A training-free alternative is model merging \citep{wortsman2022model, ilharcoediting, yadav2023ties}, which combines capabilities directly in parameter space and thereby avoids the data-level conflicts of joint optimization.
Given the shared base model $\theta_0$, the task vectors $\Delta \theta_{\text{agent}} = \theta_{\text{agent}} - \theta_0$ and $\Delta \theta_{\text{reason}} = \theta_{\text{reason}} - \theta_0$, a naive merge:
\begin{equation}
    \theta_{\text{merge}} = \theta_{\text{agent}} + \alpha \Delta \theta_{\text{reason}},
\end{equation}
where $\alpha$ is a scalar controlling the strength of the injected reasoning signal. However, operating in parameter space alone is not sufficient \cite{huang2026rain}: naive merging treats all parameter directions equally, so $\Delta \theta_{\text{reason}}$ may perturb the directions that govern the agent's think–act–observe loop, causing the merged model $\theta_{\text{merge}}$ to overthink instead of act, or to continue generating when it should seek environmental feedback.


\subsection{M2A: Behavior-Preserving Reasoning Synergy}
\label{sec:m2a}
M2A formulates the synergy between mathematical and agentic reasoning as a behavior-preserving capability integration problem: how to combine mathematical and agentic reasoning without disrupting agent’s interaction
pattern. 
M2A consists of three stages: (i) calibrating the agent-critical subspace from behavior markers, (ii) projecting the reasoning task vector onto the null space of agent-critical subspace to prevent overriding agentic behavior, and (iii) calibrating layer-wise merge coefficients and selecting compatible layers for merging.
\subsubsection{Agent-Critical Behavior Calibration}
The first stage of M2A is to identify the subspace that encodes agent-critical behaviors—the interaction patterns that a successful agent must preserve. If these behaviors are perturbed, the merged model may continue reasoning when it should act, or fail to respond appropriately to new observations.

\textbf{Behavior Markers} 
An agent's think–act–observe behavior is most clearly reflected in when it switches between internal thinking and external action: starting or stopping reasoning, making tool calls, or pausing for external feedback \cite{yao2022react, wei2026agentic}. These transitions appear in the trajectory as a few special tokens, and the activations around them carry the strongest signals of the agent's interaction behavior. We therefore identify these transitions through a set of markers: 
\begin{equation}
\mathcal{M}_{\rm native} =
\{\text{<think>}, \text{</think>},
\text{<tool\_call>}, \text{</tool\_call>}\},
\end{equation}
These markers follow our agent's chat template; other formats can be supported by redefining $\mathcal{M}$.

\textbf{Agent-critical Subspace} 
For each marker, we extract activations from its local neighborhood of $r$ tokens, chosen to capture the transition from both sides: for start markers, the activations at positions $[t_{\text{marker}}, t_{\text{marker}}+r]$ reflect the behavior after entering a new state; for end markers, the activations at positions $[t_{\text{marker}}-r, t_{\text{marker}}]$ reflect the decision to exit the current state.
For each layer $l$, we collect the corresponding input hidden states from the agent model into a calibration matrix:
\begin{equation}
\mathbf{C}^{(l)} = [\mathbf{h}_1^{(l)}, \mathbf{h}_2^{(l)}, \dots, \mathbf{h}_{n_l}^{(l)}] \in \mathbb{R}^{d \times n_l}.
\end{equation}
where $d$ is the hidden dimension and $n_l$ is the number of collected features. The subspace spanned by these calibration features:
\begin{equation}
S_{\text{agent}}^{(l)} = \text{span}(\mathbf{C}^{(l)}),
\end{equation}
captures the representation directions most relevant to the agent's think–act–observe behavior pattern. Details are provided in Appendix \ref{app:calibration_set}.


\subsubsection{Null-space Projected Merging}
The second stage is to merge the reasoning task vector into the agent model without perturbing the agent-critical subspace $S_{\text{agent}}^{(l)}$. 
Since model merging operates on every weight matrix of the LLM, for notational simplicity we use $\Delta\mathbf{W}_{\text{reason}}^{(l)}$ to denote the reasoning task vector at layer $l$ for any such matrix.
For any protected feature $\mathbf{h}_{\text{agent}} \in S_{\text{agent}}^{(l)}$, naive merging induces an undesired perturbation:
\begin{equation}
    \Delta \mathbf{h} = \Delta \mathbf{W}_{\text{reason}}^{(l)} \mathbf{h}_{\text{agent}},
\end{equation}
where $\Delta \mathbf{h}\neq0$ shifts the hidden representation away from its calibrated directions and can disrupt the agent's think–act–observe behavior.
To eliminate this perturbation, we seek a refined task vector $\widehat{\Delta \mathbf{W}}_{\text{reason}}^{(l)}$ that eliminates interference within the protected subspace  $S_{\text{agent}}^{(l)}$ while minimizing the deviation from the original reasoning behavior. This is formulated as a constrained optimization problem:
\begin{equation}
    \widehat{\Delta \mathbf{W}}_{\text{reason}}^{(l)} 
    = \arg\min_{\mathbf{W}} \| \mathbf{W} - \Delta \mathbf{W}_{\text{reason}}^{(l)} \|_F^2 
    \quad \text{s.t.} \quad 
    \mathbf{W} \mathbf{h} = \mathbf{0}, \;\; \forall \mathbf{h} \in S_{\text{agent}}^{(l)}.
\end{equation}
where $\| \cdot \|_F$ denotes the Frobenius norm.
To solve this problem, we resort to null-space projection \cite{wang2021training,huang2026rain}. 
Let $\mathbf{Q}^{(l)} \in \mathbb{R}^{d \times r_l}$ be an orthonormal basis of $S_{\text{agent}}^{(l)}$ (e.g., obtained via SVD of $\mathbf{C}^{(l)}$), where $r_l$ is the rank of $S_{\text{agent}}^{(l)}$. The projector onto the orthogonal complement of $S_{\text{agent}}^{(l)}$ is:
\begin{equation}
\mathbf{P}_{\text{null}}^{(l)} = \mathbf{I} - \mathbf{Q}^{(l)} \mathbf{Q}^{(l)\top},
\end{equation}
and the refined task vector is obtained by projecting $\Delta \mathbf{W}_{\text{reason}}^{(l)}$ along its input dimension:
\begin{equation}
    \widehat{\Delta \mathbf{W}}_{\text{reason}}^{(l)} 
    = \Delta \mathbf{W}_{\text{reason}}^{(l)} \, \mathbf{P}_{\text{null}}^{(l)}.
\end{equation}
For any $\mathbf{h} \in S_{\text{agent}}^{(l)}$, we can write $\mathbf{h} = \mathbf{Q}^{(l)} \mathbf{a}$ for coefficient vector $\mathbf{a}$. 
By the orthonormality of $\mathbf{Q}^{(l)}$:
\begin{equation}
    \widehat{\Delta \mathbf{W}}_{\text{reason}}^{(l)} \mathbf{h} 
    = \Delta \mathbf{W}_{\text{reason}}^{(l)} \mathbf{P}_{\text{null}}^{(l)} \mathbf{Q}^{(l)} \mathbf{a} 
    = \Delta \mathbf{W}_{\text{reason}}^{(l)} \left( \mathbf{Q}^{(l)} - \mathbf{Q}^{(l)} \mathbf{Q}^{(l)\top} \mathbf{Q}^{(l)} \right) \mathbf{a} 
    = \mathbf{0},
\end{equation}
which means the projected task vector introduces no change to agent-critical direction. 
Moreover, $\widehat{\Delta \mathbf{W}}_{\text{reason}}^{(l)}$ is the Frobenius-optimal solution among all matrices satisfying this constraint.
In this way, null-space projection selectively removes the components of $\Delta \mathbf{W}_{\text{reason}}^{(l)}$ that conflict with agent behavior, while keeping the rest intact for reasoning integration.
In practice, we implement this projection with an equivalent block-wise matrix-free solver to avoid explicitly materializing the full projector; see Appendix \ref{app:m2a_implementation}.

\subsubsection{Adaptive Layer-Wise Merging}
\label{sec:adaptive_transfer}
In practice, null-space projection alone is insufficient.
We find that the norms of $\Delta \theta_{\text{agent}}^{(l)}$ and $\Delta \theta_{\text{reason}}^{(l)}$ vary substantially across layers (see Figure \ref{fig:app_alpha_mask}). Consequently, a uniform merge coefficient can cause the reasoning task vector to dominate the agent's parameters, degrading its agentic behavior. 
We therefore explore the synergistic balance between internal thinking and external action by introducing two complementary mechanisms that control \textit{how much} and \textit{where} math reasoning is merged.



\paragraph{Merge Coefficient Calibration}
To address the  scale imbalance, we introduce a layer-wise calibration mechanism. We normalize the reasoning task vector to the same scale as the agentic baseline, ensuring the integration is balanced across all layers:
\begin{equation}
    \alpha_l = \beta \cdot \frac{\|\Delta \theta_{\text{agent}}^{(l)}\|_F}{\|\Delta \theta_{\text{reason}}^{(l)}\|_F + \epsilon}.
    \label{eq:alpha}
\end{equation}
where $\beta$ is a global scaling factor and $\epsilon$ is a small constant for numerical stability. This calibration removes the need to hand-tune the merge coefficient \cite{du2024parameter, yadav2023ties, yang2023adamerging} layer by layer and turns $\beta$ into a single interpretable hyperparameter: smaller $\beta$ preserves the agent's original interaction behavior, while larger $\beta$ injects stronger mathematical reasoning.

\paragraph{Similarity-aware Layer Mask}
Even after scale calibration, merging layers with conflicting updates can still destabilize the agent. 
We therefore select only those layers where the two task vectors are directionally aligned. Specifically, we compute the cosine similarity $\mathrm{sim}^{(l)}$ between $\Delta \theta_{\text{agent}}^{(l)}$ and $\Delta \theta_{\text{reason}}^{(l)}$, and retain layers whose similarity exceeds the global mean $\bar{\mathrm{sim}}$:
\begin{equation}
    M_l = \mathbb{I}\!\left[\mathrm{sim}^{(l)} > \bar{\mathrm{sim}}\right].
\end{equation}
where $M_l \in \{0,1\}$ indicates whether layer $l$ participates in the merge. 
This filters out layers where the two task vectors are less aligned, focusing the merge on regions with natural synergy and improving overall merging efficiency.


\paragraph{Final Merge Rule}
Combining null-space projection, layer-wise coefficient calibration, and similarity-aware layer mask, our final merge rule for each mergeable weight matrix at layer $l$ is:
\begin{equation}
    \mathbf{W}_{\text{merge}}^{(l)} = \mathbf{W}_{\text{agent}}^{(l)} + M_l \, \alpha_l \, \widehat{\Delta \mathbf{W}}_{\text{reason}}^{(l)}.
\end{equation}
In this way, M2A yields a merged model that synergizes mathematical and agentic reasoning: it strengthens internal reasoning while preserving the behaviorally important structure
of multi-turn interaction. The layer-wise calibration also reduces coefficient sensitivity and exposes
$\beta$ as an explicit control knob for shaping merged model behavior. Implementation details for computing $\alpha_l$ and $M_l$ from layer-wise mergeable blocks are provided in Appendix~\ref{app:adaptive_layerwise_details}.

\section{Experiments}
\subsection{Experimental Setup}

\paragraph{Baselines.}
We build our experiments on Qwen3-8B-Base \cite{yang2025qwen3} and compare our approach with two baseline families: SFT and model merging.
Regarding SFT, we employ DeepSeek-R1-0528-Qwen3-8B as the Reasoning-8B model. We also train two additional models using Step-Fun’s SFT dataset\footnote{https://huggingface.co/datasets/stepfun-ai/Step-3.5-Flash-SFT}: \textbf{Agent-8B}, trained on 30k curated coding-agent samples, and \textbf{Multi-Task-8B}, trained on a 60k mixture of coding-agent and math-reasoning data.
For model merging, we adopt five representative baselines: \textbf{Task 
  Arithmetic}~\cite{ilharcoediting}, \textbf{TIES-Merging}~\cite{yadav2023ties}, \textbf{DARE}~\cite{yu2024language}, \textbf{SLERP}~\cite{goddard2024arcee},
   and \textbf{RAIN-Merging}~\cite{huang2026rain}. We apply M2A on top of Agent-8B, and all model-merging baselines use Agent-8B and Reasoning-8B as their source models. 
   Further details are provided in  Appendix~\ref{app:experimental_details}.
\paragraph{Evaluation.}
We evaluate M2A and baselines on the challenging coding-agent benchmark SWE-Bench Verified ~\citep{jimenez2023swe} under the OpenHands scaffold \cite{openhands2025critic}, where each task requires the model to reason and interact with the environment to resolve real-world software engineering issues. 
We report the Resolved Rate as the metric of agent performance; unless otherwise specified, all reported results are averaged over three independent runs (Avg@3).
We also evaluate mathematical reasoning on AIME2024 \cite{balunovic2025matharena}, AIME2025 \cite{balunovic2025matharena}, and MATH500 \cite{hendrycks2021measuring}. Details are shown in Appendix ~\ref{app:experimental_details}.

\begin{table*}[htb]
\centering
\small
\caption{Main results on SWE-Bench Verified. Avg. Reasoning Length is the number of reasoning tokens per step, and Avg. Step represents the number of turns the model interacts with the environment. Results are averaged over 3 runs, subscripts denote standard deviation.}
\label{tab:main_results_full}
\setlength{\tabcolsep}{5.6pt}
\renewcommand{\arraystretch}{1.08}
\resizebox{\textwidth}{!}{%
\begin{tabular}{@{}l c c c c c@{}}
\toprule
\textbf{Model} & \textbf{Params} & \textbf{Training} & \textbf{Score (\%)} $\uparrow$ & \textbf{Avg.\ Reasoning Length} & \textbf{Avg.\ Step} \\
\midrule

\multicolumn{6}{c}{\textbf{Proprietary Models}} \\
\midrule
OpenAI-GPT-5.2-Thinking \cite{openai2025gpt52} & -- & -- & $80.0$ & -- & -- \\
Gemini-3.1-Pro \cite{google2026gemini31} & -- & -- & $80.6$ & -- & -- \\
Claude-Opus-4.6 \cite{anthropic2026opus46} & -- & -- & $80.8$ & -- & -- \\

\midrule

\multicolumn{6}{c}{\textbf{Open-Source Models}} \\
\midrule
SWE-agent-LM-7B \cite{yang2025swe} & 7B & SFT & $15.2$ & -- & -- \\
SWE-Mirror-LM-7B \cite{wang2025swe-mirror} & 7B & SFT & $22.8$ & -- & -- \\
SWE-Dev-7B \cite{wang2025swe} & 7B & SFT + RL & $23.4$ & -- & -- \\
Klear-Agent-8B-RL \cite{kwaiklear2025miniswe} & 8B & SFT + RL & $40.4$ & -- & -- \\
SWE-AGILE \cite{lian2026swe} & 8B & SFT + RL & $24.1$ & -- & -- \\
SWE-Lego-Qwen3-8B \cite{tao2026swe} & 8B & SFT & $42.2$ & -- & -- \\
SERA-8B \cite{shen2026sera} & 8B & SFT & $37.1$ & -- & -- \\
DeepSWE-32B-Preview \cite{luo2025deepswe} & 32B & SFT + RL & $42.2$ & -- & -- \\
SWE-Lego-Qwen3-32B \cite{tao2026swe} & 32B & SFT & $52.6$ & -- & -- \\
SWE-World-32B-RL \cite{sun2026swe} & 32B & SFT + RL & $55.0$ & -- & -- \\
Kimi-Dev-72B \cite{yang2025kimi} & 72B & SFT + RL & $48.6$ & -- & -- \\

\midrule
\multicolumn{6}{c}{\textbf{Our Implementations}} \\
\midrule
Agent-8B & 8B & SFT & $44.0_{\pm 0.9}$ & $253.3$ & $175.3$ \\
Reasoning-8B & 8B & SFT & $0.2_{\pm 0.1}$ & $\mathbf{1800.7}$ & $9.2$ \\
Multi-Task-8B & 8B & SFT & $41.1_{\pm 1.4}$ & $347.4$ & $150.1$ \\
\noalign{\smallskip}
\cdashline{1-6}[0.5pt/1.5pt]
\noalign{\smallskip}
Task Arithmetic \cite{ilharcoediting} & 8B & Training-free & $47.6_{\pm 0.9}$ & $836.9$ & $87.1$ \\
TIES-Merging \cite{yadav2023ties} & 8B & Training-free & $39.0_{\pm 0.9}$ & $494.6$ & $112.1$ \\
DARE \cite{yu2024language} & 8B & Training-free & $22.0_{\pm 0.8}$ & $1219.1$ & $68.6$ \\
SLERP \cite{goddard2024arcee} & 8B & Training-free & $47.2_{\pm 0.7}$ & $747.4$ & $87.6$ \\
RAIN-Merging \cite{huang2026rain} & 8B & Training-free & $43.2_{\pm 1.1}$ & $262.7$ & $170.4$ \\
\noalign{\smallskip}
\cdashline{1-6}[0.5pt/0.5pt]
\noalign{\smallskip}
\textbf{M2A (Ours)} & 8B & Training-free & $\mathbf{51.2_{\pm 0.6}} (\uparrow7.2)$ & $327.4$ & $\mathbf{178.0}$ \\
\bottomrule
\end{tabular}%
}
\end{table*}

\subsection{Main Results}
\label{sec:main_results}

Table \ref{tab:main_results_full} shows the results of the proposed method and baselines. To provide a more comprehensive view of our method's effectiveness, we additionally report the performance of various models and methods of different scales.
From the results, we can find that:

\textbf{Multi-task SFT struggles to transfer mathematical reasoning to agentic reasoning under data-level mismatch.}
 Multi-task SFT increases the average reasoning length from 253.3 to 347.4, yet the average number of interaction steps (Avg. Step) drops from 175.3 to 150.1 and the resolved rate decreases from
   44.0\% to 41.1\%. In other words, the longer reasoning traces do not translate into better agentic problem solving; instead, the model interacts less with the environment and solves fewer tasks.
 This suggests that jointly optimizing over mathematical and agentic data only transfers the surface pattern of long reasoning; simply mixing math reasoning and agent data is therefore insufficient for reasoning synergy.

\textbf{Naive model merging suffers from behavior pattern mismatch.}
The reasoning model and the agent model follow different behavior patterns:
the former solves problems through single-turn internal thinking, while the latter must maintain a multi-turn think--act--observe loop with the environment.
If merging methods operate directly in parameter space without distinguishing these behavior patterns, the merged model can be dominated by the single-turn reasoning behavior, empirically exhibit substantially longer reasoning length together with far fewer interaction steps.
Task Arithmetic~\cite{ilharcoediting} and SLERP~\cite{goddard2024arcee} do achieve modest gains over the base agent (47.6\% and 47.2\% vs.\ 44.0\%), but this comes at the cost of nearly halving the interaction frequency (Avg.\ Step 87.6 vs.\ 175.3) and tripling the reasoning length. 
The merged models compensate for shorter interaction with longer single-turn thinking, rather than genuinely synergizing the two behaviors, and the gain is still noticeably below M2A.
Less stable methods degrade further: TIES-Merging~\cite{yadav2023ties} drops to 39.0\% (Avg.\ Step 112.1), and DARE~\cite{yu2024language} collapses to 22.0\% with Avg.\ Step drops sharply to 68.6.
RAIN-Merging~\cite{huang2026rain} instead preserves the agent's behavior, but for the
same reason it barely injects any additional reasoning into the model, and its resolved rate (43.2\%) shows no improvement over the base agent (44.0\%).
This indicates that preserving single-turn long reasoning pattern is insufficient for agents. Effective reasoning synergy requires jointly strengthening single-turn reasoning and preserving the multi-turn interaction behavior that drives external action. We carefully tuned the hyperparameters for the baselines, and additional experimental results are presented in the Appendix \ref{app:merging_hyperparameters}.

\textbf{M2A establishes a novel paradigm for synergizing mathematical and agentic reasoning.}
By addressing the mismatches identified above, M2A improves \texttt{Agent-8B} \textbf{from 44.0\% to 51.2\%} on SWE-Bench Verified, yielding a substantial +7.2 point gain. This result is even better than several larger models trained with SFT or RL. 
Compared with Multi-task SFT, M2A avoids joint optimization over the mismatched mathematical and agentic data. 
Compared with naive model merging, M2A further addresses the behavior pattern mismatch between single-turn mathematical reasoning and multi-turn agentic interaction by protecting agent-critical behavior directions through null-space projection.
As a result, M2A strengthens internal thinking and preserves agentic interaction: Avg. reasoning length increases from 253.3 to 327.4, while Avg. Step remains high at 178.0. Consequently, M2A achieves the best agentic performance among all compared methods. These results suggest that mathematical and agentic reasoning are most effectively synergized when the integration mechanism enhances internal reasoning while preserving external action behavior. Beyond this, M2A requires no costly training, synergistically improves mathematical reasoning as well, and further offers a controllable mechanism for tuning reasoning length, as we demonstrate in later sections.

\subsection{Synergistic Effect and Ablation Study}
\label{sec:ablation}

\begin{figure}
    \centering
    \includegraphics[width=1\linewidth]{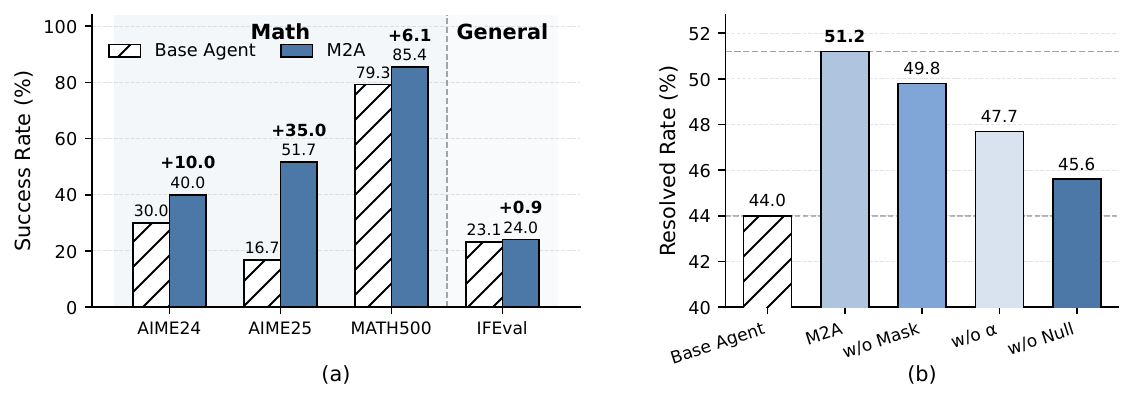}
    \caption{(a) The performance of M2A on mathematical and general capability benchmarks; (b) Ablation study of M2A on SWE-Bench Verified.}
    \label{fig:synergistic_improvement}
\end{figure}

\paragraph{Synergistic Improvement on Mathematical and Agentic Reasoning}
\label{sec:synergistic_improvement}
M2A is designed to synergize mathematical and agentic reasoning.
This design suggests that the merged model should not only improve agentic tasks, but also strengthen the mathematical reasoning capability. 
Our experiments indicate that M2A produces a \textbf{synergistic effect}: the agent becomes better at interactive problem solving while also gaining stronger intrinsic reasoning ability. As shown in Figure~\ref{fig:synergistic_improvement} (a), M2A consistently improves mathematical reasoning while also boosting SWE-Bench Verified performance. 
These gains indicate that the injected reasoning component remains effective for closed-world mathematical reasoning, while the null-space projection prevents it from overwriting the agent's behavior pattern. In addition, IFEval remains nearly unchanged, suggesting that M2A does not noticeably degrade general instruction-following ability. 
Overall, these results show that M2A enables a synergistic enhancement of mathematical and agentic reasoning.

\paragraph{Ablation Study of M2A} We ablate the three core components of M2A on SWE-Bench Verified. As shown in Figure~\ref{fig:synergistic_improvement} (b), the full method achieves the best performance, while removing any component weakens the gain.
Without Merge Coefficient Calibration (w/o $\alpha$), the merged model is no longer properly calibrated to the agent model, making the merge less effective. 
Without the Similarity-aware Layer Mask (w/o Mask), M2A cannot filter layers where the agent and reasoning task vectors are less compatible, which also reduces performance.
The most significant degradation comes from removing null-space projection (w/o Null). 
In this case, the model retains only a small improvement over the Base Agent, indicating that naive reasoning injection can interfere with the agent-critical behavior subspace. 
This validates the central mechanism of M2A: effective reasoning transfer requires preserving the agent-critical behavior, rather than simply increasing reasoning intensity.

\subsection{Merge Strength as a Control Knob for Reasoning Behavior}
\label{sec:beta_sweep}
\begin{figure}[htb]
    \centering
    \includegraphics[width=1\linewidth]{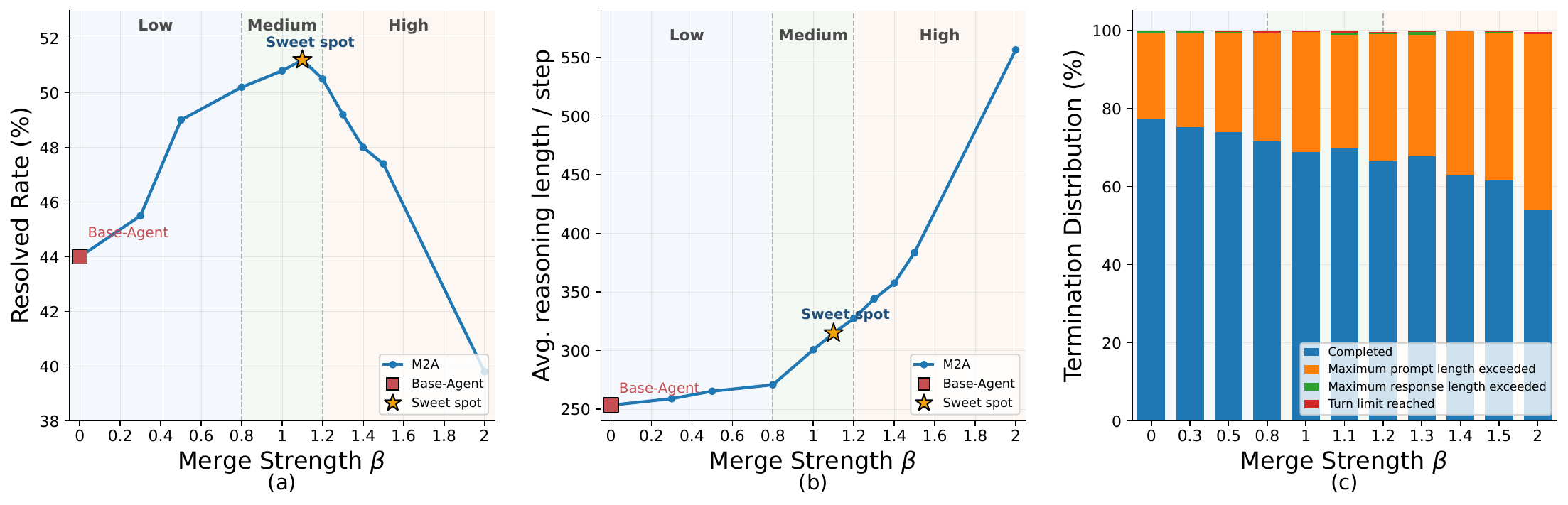}
    \caption{Effect of merge strength $\beta$ on merged model behavior. 
    (a) SWE-Bench Verified performance, (b) Avg. reasoning length per step, and (c) Termination Reason Distribution.}
    \label{fig:beta_sweep}
\end{figure}

Merge strength $\beta$ controls the intensity of the projected reasoning update injected into the agent model, and provides M2A with a \textbf{control knob} for \textbf{shaping merged model's behavior pattern}. 
As shown in Figure~\ref{fig:beta_sweep} (b), the average reasoning length per step grows approximately monotonically with $\beta$, and becomes nearly linear in the high-performing region. 
This indicates that a simple parameter-space coefficient can be translated into predictable behavior-space control over reasoning intensity. 
Meanwhile, the stable high-performing region around medium merge strength reduces the burden of manual hyperparameter search. 
We use $\beta=0.8$ and $\beta=1.2$ as empirical transition points, dividing the sweep into three behavior regimes.
This controllability distinguishes M2A from standard merging baselines. 
In Appendix~\ref{app:merging_hyperparameters}, we show that other model merging baselines do not provide a stable behavior-control interface: their reasoning length changes nonlinearly and can collapse abruptly.

In the \emph{Low Reasoning} regime, the injected reasoning signal is weak. 
The model largely preserves the behavior pattern of the original agent, but its reasoning remains weak. 
As a result, it tends to act with limited analysis and has insufficient capacity to build a reliable understanding of complex software issues before intervention.

As $\beta$ enters the \emph{Medium Reasoning} regime, the model exhibits \textbf{the most effective behavior pattern}. 
It gains stronger internal thinking while preserving the agent's ability to interact with the environment. 
This regime does not simply increase reasoning length; it shifts the agent toward more evidence-aware decision making. 
The model can spend more computation on understanding the issue, collecting relevant context, and planning edits, without drifting away from the multi-turn behavior pattern. We provide a detailed analysis of this phenomenon in the next subsection.
Importantly, performance remains \textbf{consistently high} across this interval, showing that M2A exposes \textbf{a robust behavior-control} region rather than relying on a fragile coefficient.

In the \emph{High Reasoning} regime, further increasing $\beta$ continues to lengthen reasoning traces but no longer improves performance. 
The model becomes overly dominated by internal thinking: it delays interaction, consumes more context, and is more likely to exceed its effective context-handling capacity. 
This is consistent with Figure~\ref{fig:beta_sweep}(c), where high-$\beta$ models show more failures caused by maximum prompt length exceeded.

\subsection{Trajectory-Level Analysis: From Trial-and-Error Editing to Evidence-Grounded Action}

To understand how M2A improves agent performance, we analyze trajectory-level behavior in Table~\ref{tab:trajectory_behavior_change}. We observe two consistent shifts: (i) the agent reasons more at decision-critical steps (edit and late-stage synthesis), and (ii) it performs more search and inspection before editing, while issuing fewer and more concentrated edits overall. Together, these shifts suggest that M2A changes when and how the agent commits to an action — gathering more evidence before editing rather than probing through repeated edits.
We refer to this as a shift from trial-and-error editing (frequent, exploratory edits with limited prior inspection) to evidence-grounded action (fewer edits preceded by more search and reasoning).

\begin{table}[ht]
\centering
\small
\caption{Trajectory-level changes in reasoning and action behavior after applying M2A. 
Relative change is computed as $(\mathrm{M2A}-\mathrm{Base})/\mathrm{Base}$.}
\label{tab:trajectory_behavior_change}
\setlength{\tabcolsep}{4.2pt}
\renewcommand{\arraystretch}{1.08}
\begin{tabularx}{\columnwidth}{l X c c c}
\toprule
\textbf{Metric} & \textbf{Definition} & \textbf{Base Agent} & \textbf{M2A} & \textbf{Rel. Change} \\
\midrule
Resolved Rate $\uparrow$
& Fraction of issues resolved.
& 44.0\% 
& \textbf{51.2\%} 
& \poschg{+16.4\%} \\

Reasoning at edit $\uparrow$
& Reasoning length tokens at edit steps. 
& 305.3 
& \textbf{401.4} 
& \poschg{+31.5\%} \\

Late-stage synthesis $\uparrow$
& Mean reasoning tokens near trajectory end. 
& 189.6 
& \textbf{230.7} 
& \poschg{+21.7\%} \\

Search before first edit $\uparrow$
& Search actions before the first edit. 
& 19.5 
& \textbf{25.3} 
& \poschg{+30.0\%} \\

Inspect before first edit $\uparrow$
& File inspections before the first edit. 
& 10.4 
& \textbf{12.3} 
& \poschg{+18.2\%} \\

Edit count $\downarrow$
& Edit operations per trajectory. 
& 20.1 
& \textbf{16.7} 
& \negchg{-17.0\%} \\

Unique edit files $\downarrow$
& Distinct files touched by edits.
& 6.3 
& \textbf{6.0} 
& \negchg{-5.0\%} \\
\bottomrule
\end{tabularx}
\end{table}

Regarding internal thinking, M2A strengthens reasoning at decision-critical stages. 
As shown in Table \ref{tab:trajectory_behavior_change}, the merged model reasons more at edit steps and shows stronger late-stage synthesis before finalizing a solution. 
This indicates that the additional reasoning is not merely front-loaded at the beginning of a trajectory, but remains active when the agent decides what to modify and how to complete the task. 

From the perspective of external action, M2A yields actions that are both highly selective and precisely targeted.
Before the first edit, the agent performs more search and inspection, suggesting that it gathers more evidence before intervening. Meanwhile, it uses fewer edit operations and touches slightly fewer files. 
This suggests that M2A effectively mitigates blind trial-and-error behavior and encourages the agent to localize the problem before acting.

\subsection{Generalization Across Models, Scaffolds, and Tasks}
\label{sec:generalization}

To evaluate the generalizability of M2A, we extend our
experiments beyond the original Qwen3-8B and OpenHands
setting and compare M2A against two representative
model-merging baselines. The results are summarized in
Table~\ref{tab:generalization}. Complete results with all baselines are provided in Appendix~\ref{app:additional_generalization}.

\textbf{Generalization across models and scaffolds.}
On Llama-3.1-8B \cite{grattafiori2024llama}, M2A improves the SWE-Bench Verified
score from 17.6\% to 23.0\%, outperforming baselines,
while maintaining a comparable number of interaction
steps. Without scaffold-specific retraining
or recalibration, M2A consistently achieves the highest
score 46.4\% and 49.8\%
on Mini-SWE-Agent \cite{yang2024swe} and R2E \cite{jain2025r2e}, respectively. These results demonstrate that M2A effectively generalizes across different
model families and scaffolds.

\textbf{Generalization beyond coding agents.}
On GAIA \cite{mialon2024gaia}, M2A improves the score from 73.33\% to
78.79\%, substantially outperforming baselines. Notably, M2A
achieves this improvement with a moderate increase
in reasoning length, compared with the substantially
longer reasoning traces produced by the merging
baselines. This suggests that M2A's benefits extend
beyond coding agents to general agent domain tasks.

\begin{table}[htb]
\centering
\caption{Generalization across model families, agent
scaffolds, and tasks. SWE denotes SWE-Bench Verified.}
\label{tab:generalization}

\begingroup
\footnotesize
\setlength{\tabcolsep}{5pt}
\renewcommand{\arraystretch}{0.96}

\begin{tabular*}{\linewidth}{
@{\extracolsep{\fill}}lccc@{}}
\toprule
\textbf{Method}
& \textbf{Score (\%) $\uparrow$}
& \textbf{Avg.\ Reasoning Length}
& \textbf{Avg. Step} \\
\midrule

\multicolumn{4}{@{}l}{
\textit{Llama-3.1-8B / OpenHands / SWE}} \\
Agent-8B
    & 17.6 & 207.79 & 67.66 \\
Task Arithmetic~\cite{ilharcoediting}
    & 21.0 & 483.38 & 43.73 \\
RAIN-Merging~\cite{huang2026rain}
    & 20.4 & 242.45 & 70.00 \\
\textbf{M2A}
    & \textbf{23.0} & 228.76 & 70.70 \\
\midrule
\addlinespace[4pt]
\multicolumn{4}{@{}l}{
\textit{Qwen3-8B / Mini-SWE-Agent / SWE}} \\
Agent-8B
    & 42.6 & 179.08 & 168.00 \\
Task Arithmetic~\cite{ilharcoediting}
    & 43.5 & 866.42 & 80.79 \\
RAIN-Merging~\cite{huang2026rain}
    & 35.8 & 179.40 & 189.56 \\
\textbf{M2A}
    & \textbf{46.4} & 294.97 & 169.99 \\
\midrule
\addlinespace[4pt]
\multicolumn{4}{@{}l}{
\textit{Qwen3-8B / R2E / SWE}} \\
Agent-8B
    & 46.6 & 208.93 & 149.27 \\
Task Arithmetic~\cite{ilharcoediting}
    & 48.7 & 847.47 & 70.70 \\
RAIN-Merging~\cite{huang2026rain}
    & 46.8 & 209.30 & 151.70 \\
\textbf{M2A}
    & \textbf{49.8} & 277.76 & 166.33 \\
\midrule
\addlinespace[4pt]
\multicolumn{4}{@{}l}{
\textit{Qwen3-8B / GAIA}} \\
Agent-8B
    & 73.33 & 850 & 13.65 \\
Task Arithmetic~\cite{ilharcoediting}
    & 66.1 & 2759.1 & 11.04 \\
RAIN-Merging~\cite{huang2026rain}
    & 69.1 & 2007.1 & 14.45 \\
\textbf{M2A}
    & \textbf{78.79} & 959 & 19.22 \\
\bottomrule
\end{tabular*}
\endgroup
\end{table}

\section{Conclusion}

We introduced M2A, a novel paradigm for synergizing mathematical and agentic reasoning through behavior-preserving null-space model merging. 
By protecting agent-critical behavior while injecting mathematical reasoning, M2A avoids the data-level mismatch of Multi-task SFT and the behavior pattern mismatch of naive merging. 
Experiments on SWE-Bench Verified show that M2A improves Qwen3-8B from 44.0\% to 51.2\% without additional training, while also improving mathematical reasoning and enabling controllable reasoning behavior. And experiments demonstrate that M2A
generalizes across model families, agent scaffolds, and task domains. These findings open a new direction for synergizing reasoning capabilities across fundamentally different behavior patterns.

\begin{ack}
\textbf{Funding Statement.}
This work was conducted during Junjian Wang's internship
with the Foundation Model Team at Li Auto Inc.
Financial support and computational resources for this
research were provided by Li Auto Inc.
This work was supported in part by the National Natural
Science Foundation of China (No. 62276260).

\textbf{Competing Interests.}
The authors declare no competing interests.
\end{ack}


\bibliographystyle{plainnat}
\bibliography{reference}

\begin{thebibliography}{55}
\providecommand{\natexlab}[1]{#1}
\providecommand{\url}[1]{\texttt{#1}}
\expandafter\ifx\csname urlstyle\endcsname\relax
  \providecommand{\doi}[1]{doi: #1}\else
  \providecommand{\doi}{doi: \begingroup \urlstyle{rm}\Url}\fi

\bibitem[{Anthropic}(2026)]{anthropic2026opus46}
{Anthropic}.
\newblock Introducing claude opus 4.6.
\newblock \url{https://www.anthropic.com/news/claude-opus-4-6}, February 2026.
\newblock Accessed: 2026-04-25.

\bibitem[Bai et~al.(2025)Bai, Sun, Hu, Qiu, Huan, Song, Nowak, and Song]{bai2025how}
Haoyue Bai, Yiyou Sun, Wenjie Hu, Shi Qiu, Maggie~Ziyu Huan, Peiyang Song, Robert Nowak, and Dawn Song.
\newblock How and why llms generalize: A fine-grained analysis of llm reasoning from cognitive behaviors to low-level patterns.
\newblock \emph{arXiv preprint arXiv:2512.24063}, 2025.

\bibitem[Balunovi{\'c} et~al.(2025)Balunovi{\'c}, Dekoninck, Petrov, Jovanovi{\'c}, and Vechev]{balunovic2025matharena}
Mislav Balunovi{\'c}, Jasper Dekoninck, Ivo Petrov, Nikola Jovanovi{\'c}, and Martin Vechev.
\newblock Matharena: Evaluating llms on uncontaminated math competitions.
\newblock \emph{arXiv preprint arXiv:2505.23281}, 2025.

\bibitem[Comanici et~al.(2025)Comanici, Bieber, Schaekermann, Pasupat, Sachdeva, Dhillon, Blistein, Ram, Zhang, Rosen, et~al.]{comanici2025gemini}
Gheorghe Comanici, Eric Bieber, Mike Schaekermann, Ice Pasupat, Noveen Sachdeva, Inderjit Dhillon, Marcel Blistein, Ori Ram, Dan Zhang, Evan Rosen, et~al.
\newblock Gemini 2.5: Pushing the frontier with advanced reasoning, multimodality, long context, and next generation agentic capabilities.
\newblock \emph{arXiv preprint arXiv:2507.06261}, 2025.

\bibitem[Du et~al.(2024)Du, Lee, Li, Jiang, Guo, Yu, Liu, Goh, Tang, He, et~al.]{du2024parameter}
Guodong Du, Junlin Lee, Jing Li, Runhua Jiang, Yifei Guo, Shuyang Yu, Hanting Liu, Sim~K Goh, Ho-Kin Tang, Daojing He, et~al.
\newblock Parameter competition balancing for model merging.
\newblock \emph{Advances in Neural Information Processing Systems}, 37:\penalty0 84746--84776, 2024.

\bibitem[Fang et~al.(2024)Fang, Jiang, Wang, Ma, Jie, Wang, He, and Chua]{fang2024alphaedit}
Junfeng Fang, Houcheng Jiang, Kun Wang, Yunshan Ma, Shi Jie, Xiang Wang, Xiangnan He, and Tat-Seng Chua.
\newblock Alphaedit: Null-space constrained knowledge editing for language models.
\newblock \emph{arXiv preprint arXiv:2410.02355}, 2024.

\bibitem[Goddard et~al.(2024)Goddard, Siriwardhana, Ehghaghi, Meyers, Karpukhin, Benedict, McQuade, and Solawetz]{goddard2024arcee}
Charles Goddard, Shamane Siriwardhana, Malikeh Ehghaghi, Luke Meyers, Vladimir Karpukhin, Brian Benedict, Mark McQuade, and Jacob Solawetz.
\newblock Arcee’s mergekit: A toolkit for merging large language models.
\newblock In \emph{Proceedings of the 2024 Conference on Empirical Methods in Natural Language Processing: Industry Track}, pages 477--485, 2024.

\bibitem[{Google DeepMind}(2026)]{google2026gemini31}
{Google DeepMind}.
\newblock Gemini 3.1 pro.
\newblock \url{https://deepmind.google/models/gemini/pro/}, 2026.
\newblock Accessed: 2026-04-25.

\bibitem[Grattafiori et~al.(2024)Grattafiori, Dubey, Jauhri, Pandey, Kadian, Al-Dahle, Letman, Mathur, Schelten, Vaughan, et~al.]{grattafiori2024llama}
Aaron Grattafiori, Abhimanyu Dubey, Abhinav Jauhri, Abhinav Pandey, Abhishek Kadian, Ahmad Al-Dahle, Aiesha Letman, Akhil Mathur, Alan Schelten, Alex Vaughan, et~al.
\newblock The llama 3 herd of models.
\newblock \emph{arXiv preprint arXiv:2407.21783}, 2024.

\bibitem[Guo et~al.(2025)Guo, Yang, Zhang, Song, Wang, Zhu, Xu, Zhang, Ma, Bi, et~al.]{guo2025deepseek}
Daya Guo, Dejian Yang, Haowei Zhang, Junxiao Song, Peiyi Wang, Qihao Zhu, Runxin Xu, Ruoyu Zhang, Shirong Ma, Xiao Bi, et~al.
\newblock Deepseek-r1: Incentivizing reasoning capability in llms via reinforcement learning.
\newblock \emph{arXiv preprint arXiv:2501.12948}, 2025.

\bibitem[Hendrycks et~al.(2021)Hendrycks, Burns, Kadavath, Arora, Basart, Tang, Song, and Steinhardt]{hendrycks2021measuring}
Dan Hendrycks, Collin Burns, Saurav Kadavath, Akul Arora, Steven Basart, Eric Tang, Dawn Song, and Jacob Steinhardt.
\newblock Measuring mathematical problem solving with the math dataset.
\newblock \emph{arXiv preprint arXiv:2103.03874}, 2021.

\bibitem[Huan et~al.(2025)Huan, Li, Zheng, Xu, Kim, Du, Poovendran, Neubig, and Yue]{huan2025does}
Maggie Huan, Yuetai Li, Tuney Zheng, Xiaoyu Xu, Seungone Kim, Minxin Du, Radha Poovendran, Graham Neubig, and Xiang Yue.
\newblock Does math reasoning improve general llm capabilities? understanding transferability of llm reasoning.
\newblock \emph{arXiv preprint arXiv:2507.00432}, 2025.

\bibitem[Huang et~al.(2026)Huang, Liu, Lin, Lou, He, Tian, Li, and Huang]{huang2026rain}
Zhehao Huang, Yuhang Liu, Baijiong Lin, Yixin Lou, Zhengbao He, Hanling Tian, Tao Li, and Xiaolin Huang.
\newblock Rain-merging: A gradient-free method to enhance instruction following in large reasoning models with preserved thinking format.
\newblock \emph{arXiv preprint arXiv:2602.22538}, 2026.

\bibitem[Ilharco et~al.()Ilharco, Ribeiro, Wortsman, Schmidt, Hajishirzi, and Farhadi]{ilharcoediting}
Gabriel Ilharco, Marco~Tulio Ribeiro, Mitchell Wortsman, Ludwig Schmidt, Hannaneh Hajishirzi, and Ali Farhadi.
\newblock Editing models with task arithmetic.
\newblock In \emph{The Eleventh International Conference on Learning Representations}.

\bibitem[Jaech et~al.(2024)Jaech, Kalai, Lerer, Richardson, El-Kishky, Low, Helyar, Madry, Beutel, Carney, et~al.]{jaech2024openai}
Aaron Jaech, Adam Kalai, Adam Lerer, Adam Richardson, Ahmed El-Kishky, Aiden Low, Alec Helyar, Aleksander Madry, Alex Beutel, Alex Carney, et~al.
\newblock Openai o1 system card.
\newblock \emph{arXiv preprint arXiv:2412.16720}, 2024.

\bibitem[Jain et~al.(2025)Jain, Singh, Shetty, Zheng, Sen, and Stoica]{jain2025r2e}
Naman Jain, Jaskirat Singh, Manish Shetty, Liang Zheng, Koushik Sen, and Ion Stoica.
\newblock R2e-gym: Procedural environments and hybrid verifiers for scaling open-weights swe agents.
\newblock \emph{arXiv preprint arXiv:2504.07164}, 2025.

\bibitem[Jimenez et~al.(2023)Jimenez, Yang, Wettig, Yao, Pei, Press, and Narasimhan]{jimenez2023swe}
Carlos~E Jimenez, John Yang, Alexander Wettig, Shunyu Yao, Kexin Pei, Ofir Press, and Karthik Narasimhan.
\newblock Swe-bench: Can language models resolve real-world github issues?
\newblock \emph{arXiv preprint arXiv:2310.06770}, 2023.

\bibitem[Kojima et~al.(2022)Kojima, Gu, Reid, Matsuo, and Iwasawa]{kojima2022large}
Takeshi Kojima, Shixiang~Shane Gu, Machel Reid, Yutaka Matsuo, and Yusuke Iwasawa.
\newblock Large language models are zero-shot reasoners.
\newblock \emph{Advances in neural information processing systems}, 35:\penalty0 22199--22213, 2022.

\bibitem[{Kwai-Klear}(2025)]{kwaiklear2025miniswe}
{Kwai-Klear}.
\newblock mini-swe-agent-plus: The 100-line {AI} agent that solves {GitHub} issues with text-edit tool.
\newblock \url{https://github.com/Kwai-Klear/mini-swe-agent-plus}, 2025.
\newblock GitHub Repository.

\bibitem[Li et~al.(2026)Li, Yi, Li, Fan, Jiang, Chen, Li, Song, and Zhang]{li2026reasoning}
Yu~Li, Mingyang Yi, Xiuyu Li, Ju~Fan, Fuxin Jiang, Binbin Chen, Peng Li, Jie Song, and Tieying Zhang.
\newblock Reasoning and tool-use compete in agentic rl: From quantifying interference to disentangled tuning.
\newblock \emph{arXiv preprint arXiv:2602.00994}, 2026.

\bibitem[Lian et~al.(2026)Lian, Liu, Chen, Chen, and Li]{lian2026swe}
Shuquan Lian, Juncheng Liu, Yazhe Chen, Yuhong Chen, and Hui Li.
\newblock Swe-agile: A software agent framework for efficiently managing dynamic reasoning context.
\newblock \emph{arXiv preprint arXiv:2604.11716}, 2026.

\bibitem[Luo et~al.(2025)Luo, Jain, Singh, Tan, Patel, Wu, Ariyak, Cai, Venkat, Zhu, Athiwaratkun, Roongta, Zhang, Li, Popa, Sen, and Stoica]{luo2025deepswe}
M.~Luo, N.~Jain, J.~Singh, S.~Tan, A.~Patel, Q.~Wu, A.~Ariyak, C.~Cai, T.~Venkat, S.~Zhu, B.~Athiwaratkun, M.~Roongta, C.~Zhang, L.~E. Li, R.~A. Popa, K.~Sen, and I.~Stoica.
\newblock {DeepSWE}: Training a fully open-sourced, state-of-the-art coding agent by scaling {RL}.
\newblock \url{https://www.together.ai/blog/deepswe}, July 2025.
\newblock Together AI Blog post.

\bibitem[Ma et~al.(2025{\natexlab{a}})Ma, Liu, Chen, Zhang, and Shao]{ma2025led}
Qianli Ma, Dongrui Liu, Qian Chen, Linfeng Zhang, and Jing Shao.
\newblock Led-merging: Mitigating safety-utility conflicts in model merging with location-election-disjoint.
\newblock In \emph{Proceedings of the 63rd Annual Meeting of the Association for Computational Linguistics (Volume 1: Long Papers)}, pages 21749--21767, 2025{\natexlab{a}}.

\bibitem[Ma et~al.(2025{\natexlab{b}})Ma, Liu, Jiang, Zhang, Ma, and Chen]{ma2025general}
Xueguang Ma, Qian Liu, Dongfu Jiang, Ge~Zhang, Zejun Ma, and Wenhu Chen.
\newblock General-reasoner: Advancing llm reasoning across all domains.
\newblock \emph{arXiv preprint arXiv:2505.14652}, 2025{\natexlab{b}}.

\bibitem[Mialon et~al.(2024)Mialon, Fourrier, Wolf, LeCun, and Scialom]{mialon2024gaia}
Gr{\'e}goire Mialon, Cl{\'e}mentine Fourrier, Thomas Wolf, Yann LeCun, and Thomas Scialom.
\newblock Gaia: a benchmark for general ai assistants.
\newblock In \emph{International Conference on Learning Representations}, volume 2024, pages 9025--9049, 2024.

\bibitem[{OpenAI}(2025)]{openai2025gpt52}
{OpenAI}.
\newblock Introducing gpt-5.2.
\newblock \url{https://openai.com/index/introducing-gpt-5-2/}, December 2025.
\newblock Accessed: 2026-04-25.

\bibitem[{OpenHands Team}(2025)]{openhands2025critic}
{OpenHands Team}.
\newblock {OpenHands} critic 32b exp 20250417.
\newblock \url{https://huggingface.co/OpenHands/openhands-critic-32b-exp-20250417}, 2025.
\newblock Accessed: 2025-12-08.

\bibitem[Pang et~al.(2025)Pang, Kong, Savarese, Xiong, and Zhou]{pang2025reasoning}
Bo~Pang, Deqian Kong, Silvio Savarese, Caiming Xiong, and Yingbo Zhou.
\newblock Reasoning curriculum: Bootstrapping broad llm reasoning from math.
\newblock \emph{arXiv preprint arXiv:2510.26143}, 2025.

\bibitem[Qi et~al.(2025)Qi, Li, Wang, Gao, Li, Ye, and Zhou]{qi2025less}
Biqing Qi, Fangyuan Li, Zhen Wang, Junqi Gao, Dong Li, Peng Ye, and Bowen Zhou.
\newblock Less is more: Efficient model merging with binary task switch.
\newblock In \emph{Proceedings of the Computer Vision and Pattern Recognition Conference}, pages 15265--15274, 2025.

\bibitem[Ramesh et~al.(2026)Ramesh, Ji, Zimmer, Yoon, Wang, Ammar, Lucchi, and Bogunovic]{ramesh2026multitask}
Shyam~Sundhar Ramesh, Xiaotong Ji, Matthieu Zimmer, Sangwoong Yoon, Zhiyong Wang, Haitham~Bou Ammar, Aurelien Lucchi, and Ilija Bogunovic.
\newblock Multi-task grpo: Reliable llm reasoning across tasks.
\newblock \emph{arXiv preprint arXiv:2602.05547}, 2026.

\bibitem[Shen et~al.(2026)Shen, Tormoen, Shah, Farhadi, and Dettmers]{shen2026sera}
Ethan Shen, Danny Tormoen, Saurabh Shah, Ali Farhadi, and Tim Dettmers.
\newblock Sera: Soft-verified efficient repository agents.
\newblock \emph{arXiv preprint arXiv:2601.20789}, 2026.

\bibitem[Shinn et~al.(2023)Shinn, Cassano, Gopinath, Narasimhan, and Yao]{shinn2023reflexion}
Noah Shinn, Federico Cassano, Ashwin Gopinath, Karthik Narasimhan, and Shunyu Yao.
\newblock Reflexion: Language agents with verbal reinforcement learning.
\newblock \emph{Advances in neural information processing systems}, 36:\penalty0 8634--8652, 2023.

\bibitem[Su et~al.(2025)Su, Yu, Song, Li, Mi, Tu, Zhang, and Yu]{su2025crossing}
Yi~Su, Dian Yu, Linfeng Song, Juntao Li, Haitao Mi, Zhaopeng Tu, Min Zhang, and Dong Yu.
\newblock Crossing the reward bridge: Expanding rl with verifiable rewards across diverse domains.
\newblock \emph{arXiv preprint arXiv:2503.23829}, 2025.

\bibitem[Sun et~al.(2026)Sun, Song, Huang, Jiang, Le, Lv, Chen, Hu, Luo, Zhao, et~al.]{sun2026swe}
Shuang Sun, Huatong Song, Lisheng Huang, Jinhao Jiang, Ran Le, Zhihao Lv, Zongchao Chen, Yiwen Hu, Wenyang Luo, Wayne~Xin Zhao, et~al.
\newblock Swe-world: Building software engineering agents in docker-free environments.
\newblock \emph{arXiv preprint arXiv:2602.03419}, 2026.

\bibitem[Tao et~al.(2026)Tao, Chen, Jiang, Kou, Wang, Wang, Li, Yang, Du, Dai, et~al.]{tao2026swe}
Chaofan Tao, Jierun Chen, Yuxin Jiang, Kaiqi Kou, Shaowei Wang, Ruoyu Wang, Xiaohui Li, Sidi Yang, Yiming Du, Jianbo Dai, et~al.
\newblock Swe-lego: Pushing the limits of supervised fine-tuning for software issue resolving.
\newblock \emph{arXiv preprint arXiv:2601.01426}, 2026.

\bibitem[Wang et~al.(2026)Wang, Long, Li, Xu, Li, and Tang]{wang2026mix}
Haoqing Wang, Xiang Long, Ziheng Li, Yilong Xu, Tingguang Li, and Yehui Tang.
\newblock To mix or to merge: Toward multi-domain reinforcement learning for large language models.
\newblock \emph{arXiv preprint arXiv:2602.12566}, 2026.

\bibitem[Wang et~al.(2025{\natexlab{a}})Wang, Hou, Wei, Tang, and Dong]{wang2025swe}
Haoran Wang, Zhenyu Hou, Yao Wei, Jie Tang, and Yuxiao Dong.
\newblock Swe-dev: Building software engineering agents with training and inference scaling.
\newblock In \emph{Findings of the Association for Computational Linguistics: ACL 2025}, pages 3742--3761, 2025{\natexlab{a}}.

\bibitem[Wang et~al.(2025{\natexlab{b}})Wang, Zan, Xin, Liu, Wu, and Shen]{wang2025swe-mirror}
Junhao Wang, Daoguang Zan, Shulin Xin, Siyao Liu, Yurong Wu, and Kai Shen.
\newblock Swe-mirror: Scaling issue-resolving datasets by mirroring issues across repositories.
\newblock \emph{arXiv preprint arXiv:2509.08724}, 2025{\natexlab{b}}.

\bibitem[Wang et~al.(2021)Wang, Li, Sun, and Xu]{wang2021training}
Shipeng Wang, Xiaorong Li, Jian Sun, and Zongben Xu.
\newblock Training networks in null space of feature covariance for continual learning.
\newblock In \emph{Proceedings of the IEEE/CVF conference on Computer Vision and Pattern Recognition}, pages 184--193, 2021.

\bibitem[Wei et~al.(2022)Wei, Wang, Schuurmans, Bosma, Xia, Chi, Le, Zhou, et~al.]{wei2022chain}
Jason Wei, Xuezhi Wang, Dale Schuurmans, Maarten Bosma, Fei Xia, Ed~Chi, Quoc~V Le, Denny Zhou, et~al.
\newblock Chain-of-thought prompting elicits reasoning in large language models.
\newblock \emph{Advances in neural information processing systems}, 35:\penalty0 24824--24837, 2022.

\bibitem[Wei et~al.(2026)Wei, Li, Liu, Ning, Yang, Zou, Zeng, Qiu, Lin, Fu, et~al.]{wei2026agentic}
Tianxin Wei, Ting-Wei Li, Zhining Liu, Xuying Ning, Ze~Yang, Jiaru Zou, Zhichen Zeng, Ruizhong Qiu, Xiao Lin, Dongqi Fu, et~al.
\newblock Agentic reasoning for large language models.
\newblock \emph{arXiv preprint arXiv:2601.12538}, 2026.

\bibitem[Wortsman et~al.(2022)Wortsman, Ilharco, Gadre, Roelofs, Gontijo-Lopes, Morcos, Namkoong, Farhadi, Carmon, Kornblith, et~al.]{wortsman2022model}
Mitchell Wortsman, Gabriel Ilharco, Samir~Ya Gadre, Rebecca Roelofs, Raphael Gontijo-Lopes, Ari~S Morcos, Hongseok Namkoong, Ali Farhadi, Yair Carmon, Simon Kornblith, et~al.
\newblock Model soups: averaging weights of multiple fine-tuned models improves accuracy without increasing inference time.
\newblock In \emph{International conference on machine learning}, pages 23965--23998. PMLR, 2022.

\bibitem[Yadav et~al.(2023)Yadav, Tam, Choshen, Raffel, and Bansal]{yadav2023ties}
Prateek Yadav, Derek Tam, Leshem Choshen, Colin~A Raffel, and Mohit Bansal.
\newblock Ties-merging: Resolving interference when merging models.
\newblock \emph{Advances in neural information processing systems}, 36:\penalty0 7093--7115, 2023.

\bibitem[Yang et~al.(2025{\natexlab{a}})Yang, Li, Yang, Zhang, Hui, Zheng, Yu, Gao, Huang, Lv, et~al.]{yang2025qwen3}
An~Yang, Anfeng Li, Baosong Yang, Beichen Zhang, Binyuan Hui, Bo~Zheng, Bowen Yu, Chang Gao, Chengen Huang, Chenxu Lv, et~al.
\newblock Qwen3 technical report.
\newblock \emph{arXiv preprint arXiv:2505.09388}, 2025{\natexlab{a}}.

\bibitem[Yang et~al.(2023)Yang, Wang, Shen, Liu, Guo, Wang, and Tao]{yang2023adamerging}
Enneng Yang, Zhenyi Wang, Li~Shen, Shiwei Liu, Guibing Guo, Xingwei Wang, and Dacheng Tao.
\newblock Adamerging: Adaptive model merging for multi-task learning.
\newblock \emph{arXiv preprint arXiv:2310.02575}, 2023.

\bibitem[Yang et~al.(2026{\natexlab{a}})Yang, Shen, Guo, Wang, Cao, Zhang, and Tao]{yang2026model}
Enneng Yang, Li~Shen, Guibing Guo, Xingwei Wang, Xiaochun Cao, Jie Zhang, and Dacheng Tao.
\newblock Model merging in llms, mllms, and beyond: Methods, theories, applications, and opportunities.
\newblock \emph{ACM Computing Surveys}, 58\penalty0 (8):\penalty0 1--41, 2026{\natexlab{a}}.

\bibitem[Yang et~al.(2024)Yang, Jimenez, Wettig, Lieret, Yao, Narasimhan, and Press]{yang2024swe}
John Yang, Carlos Jimenez, Alexander Wettig, Kilian Lieret, Shunyu Yao, Karthik Narasimhan, and Ofir Press.
\newblock Swe-agent: Agent-computer interfaces enable automated software engineering.
\newblock \emph{Advances in Neural Information Processing Systems}, 37:\penalty0 50528--50652, 2024.

\bibitem[Yang et~al.(2025{\natexlab{b}})Yang, Lieret, Jimenez, Wettig, Khandpur, Zhang, Hui, Press, Schmidt, and Yang]{yang2025swe}
John Yang, Kilian Lieret, Carlos~E Jimenez, Alexander Wettig, Kabir Khandpur, Yanzhe Zhang, Binyuan Hui, Ofir Press, Ludwig Schmidt, and Diyi Yang.
\newblock Swe-smith: Scaling data for software engineering agents.
\newblock \emph{arXiv preprint arXiv:2504.21798}, 2025{\natexlab{b}}.

\bibitem[Yang et~al.(2026{\natexlab{b}})Yang, Shi, and Liu]{yang2026orthogonal}
Sihan Yang, Kexuan Shi, and Weiyang Liu.
\newblock Orthogonal model merging.
\newblock \emph{arXiv preprint arXiv:2602.05943}, 2026{\natexlab{b}}.

\bibitem[Yang et~al.(2025{\natexlab{c}})Yang, Wang, Fu, He, Xiong, Liu, Miao, Gao, Wang, Ma, et~al.]{yang2025kimi}
Zonghan Yang, Shengjie Wang, Kelin Fu, Wenyang He, Weimin Xiong, Yibo Liu, Yibo Miao, Bofei Gao, Yejie Wang, Yingwei Ma, et~al.
\newblock Kimi-dev: Agentless training as skill prior for swe-agents.
\newblock \emph{arXiv preprint arXiv:2509.23045}, 2025{\natexlab{c}}.

\bibitem[Yao et~al.(2022)Yao, Zhao, Yu, Du, Shafran, Narasimhan, and Cao]{yao2022react}
Shunyu Yao, Jeffrey Zhao, Dian Yu, Nan Du, Izhak Shafran, Karthik~R Narasimhan, and Yuan Cao.
\newblock React: Synergizing reasoning and acting in language models.
\newblock In \emph{The eleventh international conference on learning representations}, 2022.

\bibitem[Yu et~al.(2024)Yu, Yu, Yu, Huang, and Li]{yu2024language}
Le~Yu, Bowen Yu, Haiyang Yu, Fei Huang, and Yongbin Li.
\newblock Language models are super mario: Absorbing abilities from homologous models as a free lunch.
\newblock In \emph{Forty-first International Conference on Machine Learning}, 2024.

\bibitem[Zhang and Zhou(2025)]{zhang2025unraveling}
Haobo Zhang and Jiayu Zhou.
\newblock Unraveling lora interference: Orthogonal subspaces for robust model merging.
\newblock In \emph{Proceedings of the 63rd Annual Meeting of the Association for Computational Linguistics (Volume 1: Long Papers)}, pages 26459--26472, 2025.

\bibitem[Zhao et~al.(2026)Zhao, Chen, Meng, Li, Chen, Xu, Sun, Zhao, Song, Zhang, et~al.]{zhao2026immersion}
Jiale Zhao, Guoxin Chen, Fanzhe Meng, Minghao Li, Jie Chen, Hui Xu, Yongshuai Sun, Wayne~Xin Zhao, Ruihua Song, Yuan Zhang, et~al.
\newblock Immersion in the github universe: Scaling coding agents to mastery.
\newblock \emph{arXiv preprint arXiv:2602.09892}, 2026.

\bibitem[Zheng et~al.(2025)Zheng, Wang, Huang, Wang, Chen, Fan, Hu, and Ye]{zheng2025decouple}
Shenghe Zheng, Hongzhi Wang, Chenyu Huang, Xiaohui Wang, Tao Chen, Jiayuan Fan, Shuyue Hu, and Peng Ye.
\newblock Decouple and orthogonalize: A data-free framework for lora merging.
\newblock \emph{arXiv preprint arXiv:2505.15875}, 2025.

\end{thebibliography}
\newpage
\appendix
\section*{Appendix}

\section{Additional Experimental Details}
\label{app:experimental_details}

\subsection{SWE-Bench Verified Evaluation Setup}
\label{app:swe_eval_setup}

\paragraph{Evaluation Dataset.}
We primarily evaluate agentic reasoning on SWE-Bench Verified~\citep{jimenez2023swe}, a human-validated subset of SWE-Bench consisting of 500 solvable instances curated from real-world GitHub issues. 
The benchmark spans 12 open-source Python repositories and is designed to assess the end-to-end issue resolution capability of LLM-based agents. 
Each instance provides a repository state and an issue description, and the agent is required to inspect the codebase, interact with the environment, modify relevant files, and submit a patch. 
A task is considered successfully resolved only if the submitted patch passes the official evaluation tests.

\paragraph{Evaluation Metric.}
We report the resolved rate as Avg@3 on SWE-Bench Verified.
For each model, we run three independent trajectories under the same evaluation configuration and report the average resolved rate across the three runs.
We also report Avg. reasoning length, computed as the average number of tokens in the model-generated chain-of-thought content, i.e., the \texttt{<think>} $\cdots$ \texttt{</think>} block, over all steps.

\paragraph{Evaluation Environment.}
We use a self-hosted environment based on OpenHands. 
Specifically, all models are evaluated with the openhands scaffold and the Docker backend. 
\textbf{To prevent reward hacking} through repository history, we remove historical git logs from the evaluation repositories before running the agent. 
This ensures that the model cannot exploit commands such as inspecting previous commits to obtain solution-relevant information, and must instead rely on code understanding, environment interaction, and patch generation.

During SWE-Bench Verified evaluation, we use 8 $\times$ H200 GPUs. 
At inference time, YaRN RoPE scaling is set to 8.0, and the maximum context length is set to 262k tokens. 
The detailed configuration is shown in Table~\ref{tab:swe_eval_config}.

\begin{table}[h]
\centering
\small
\caption{\textbf{SWE-Bench Verified evaluation configuration.} All models are evaluated with the same self-hosted OpenHands scaffold and Docker backend.}
\label{tab:swe_eval_config}
\begin{tabular}{ll}
\toprule
\textbf{Item} & \textbf{Configuration} \\
\midrule
Benchmark & SWE-Bench Verified \\
Metric & resolved rate (Avg@3) \\
Scaffold & \texttt{openhands} \\
Environment & Self-hosted environment \\
Backend & \texttt{docker} \\
Git history & Removed before evaluation \\
Evaluation GPUs & 8 $\times$ H200 \\
RoPE scaling at inference & YaRN, factor = 8.0 \\
Maximum context length & 262144 \\
Maximum steps & 400 \\
Step timeout & 600 s \\
Trajectory timeout & 14400 s \\
Reward timeout & 1200 s \\
Maximum prompt length & 262144 \\
Maximum response length & 81920 \\
Temperature & 0.7 \\
Top-$p$ & 1.0 \\
Tool-call format & \texttt{native} \\
\bottomrule
\end{tabular}
\end{table}

\subsection{SFT Training Details}
\label{app:sft_details}

Table~\ref{tab:sft_details} reports the supervised fine-tuning settings for Agent-8B and Multi-Task-8B. 
Both models are trained with \texttt{ms-swift} from the same base model, \texttt{Qwen3-8B-Base}. To better align the training format with the ReAct-style think--act pattern, we process code-agent trajectories into an explicit reasoning-and-action format:
\texttt{<think>} reasoning and response \texttt{</think>} + \texttt{<tool\_call>} $\cdots$ \texttt{</tool\_call>}.
For long-context training, we use YaRN RoPE scaling with \texttt{factor=4.0} and \texttt{type=yarn}.

\begin{table}[h]
\centering
\small
\begin{threeparttable}
\caption{\textbf{SFT training settings.} Agent-8B is trained only on code-agent trajectories, while Multi-Task-8B is trained on a mixture of code-agent and mathematical reasoning trajectories.}
\label{tab:sft_details}

\begin{tabularx}{\textwidth}{lXX}
\toprule
\textbf{Setting} & \textbf{Agent-8B} & \textbf{Multi-Task-8B} \\
\midrule
Base model 
& \texttt{Qwen3-8B-Base} 
& \texttt{Qwen3-8B-Base} \\

Training data 
& 30k code-agent trajectories from Step-3.5-Flash-SFT\tnote{1}
& 30k code-agent trajectories and 30k mathematical reasoning trajectories from Step-3.5-Flash-SFT\tnote{1} \\

Training framework 
& \texttt{ms-swift}\tnote{2} 
& \texttt{ms-swift}\tnote{2} \\

Optimizer 
& Adam 
& Adam \\

Learning rate 
& $5\times 10^{-5}$ 
& $5\times 10^{-5}$ \\

Epochs 
& 2 
& 3 \\

Global batch size 
& 64 
& 64 \\

Cutoff length 
& 131072 
& 131072 \\

RoPE scaling 
& YaRN, factor = 4.0, type = yarn
& YaRN, factor = 4.0, type = yarn \\
\bottomrule
\end{tabularx}

\begin{tablenotes}[flushleft]
\footnotesize
\item[1] \url{https://huggingface.co/datasets/stepfun-ai/Step-3.5-Flash-SFT}
\end{tablenotes}

\begin{tablenotes}[flushleft]
\footnotesize
\item[2] \url{https://github.com/modelscope/ms-swift.git}
\end{tablenotes}

\end{threeparttable}
\end{table}

We verify that all training data used in this work have no instance-level overlap with SWE-Bench Verified. 
Specifically, the code-agent trajectories used to train Agent-8B and Multi-Task-8B do not contain any of the 500 SWE-Bench Verified evaluation instances. 
Therefore, the reported SWE-Bench Verified results are not affected by training-data leakage from the evaluation set.

\subsubsection{Epoch Sweep for Multi-Task SFT}
\label{app:multitask_epoch_sweep}

To rule out the possibility that the weak performance of Multi-Task-8B is simply caused by an under-tuned training schedule, we conduct an epoch sweep while keeping the training data mixture, base model, training framework, and other optimization settings unchanged. 
As shown in Table~\ref{tab:multitask_epoch_sweep}, increasing the number of epochs does not lead to a consistent improvement on SWE-Bench Verified. 
Multi-Task-8B obtains a resolved rate of 40.0 at 2 epochs, 41.1 at 3 epochs, 40.1 at 4 epochs, and 39.2 at 5 epochs. 
These results indicate that the limitation of multi-task SFT is not insufficient training, but rather the data-level mismatch introduced by jointly optimizing heterogeneous mathematical reasoning and multi-turn agent trajectories.

\begin{table}[h]
\centering
\small
\caption{\textbf{Epoch sweep for Multi-Task SFT on SWE-Bench Verified.} 
We keep the data mixture and optimization settings fixed, and only vary the number of training epochs. 
Increasing training epochs does not effectively improve agent performance.}
\label{tab:multitask_epoch_sweep}
\begin{tabular}{cc}
\toprule
\textbf{Training Epochs} & \textbf{Resolved Rate} \\
\midrule
2 & 40.0 \\
3 & 41.1 \\
4 & 40.1 \\
5 & 39.2 \\
\bottomrule
\end{tabular}
\end{table}

\subsection{Merging Hyperparameters}
\label{app:merging_hyperparameters}









\begin{figure}[htb]
    \centering
    \includegraphics[width=1.0\linewidth]{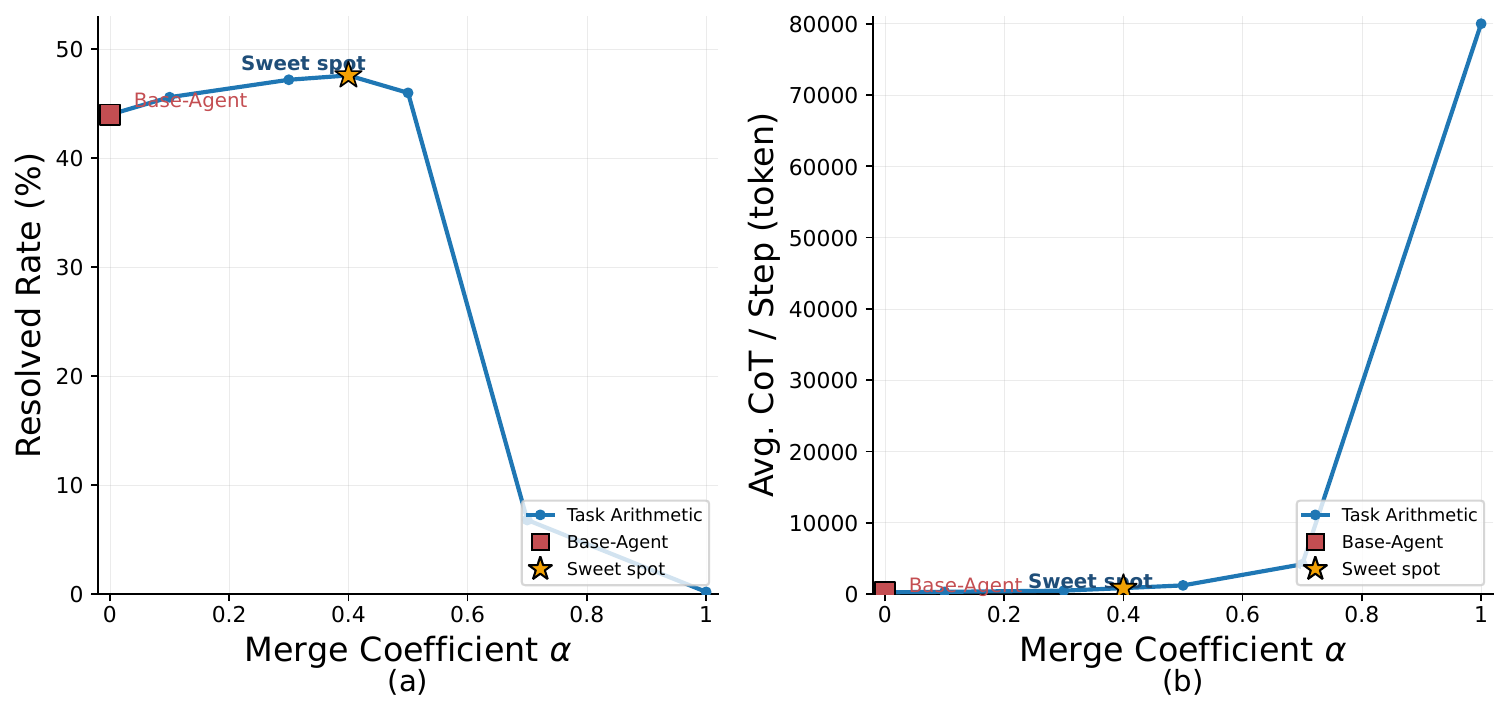}
    \caption{The effect of merge coefficient for Task Arithmetic. (a) SWE-Bench Verified performance, (b) Avg.CoT length per step}
    \label{fig:task_arithmetic_alpha_sweep}
\end{figure}

\begin{figure}[htb]
    \centering
    \includegraphics[width=1.0\linewidth]{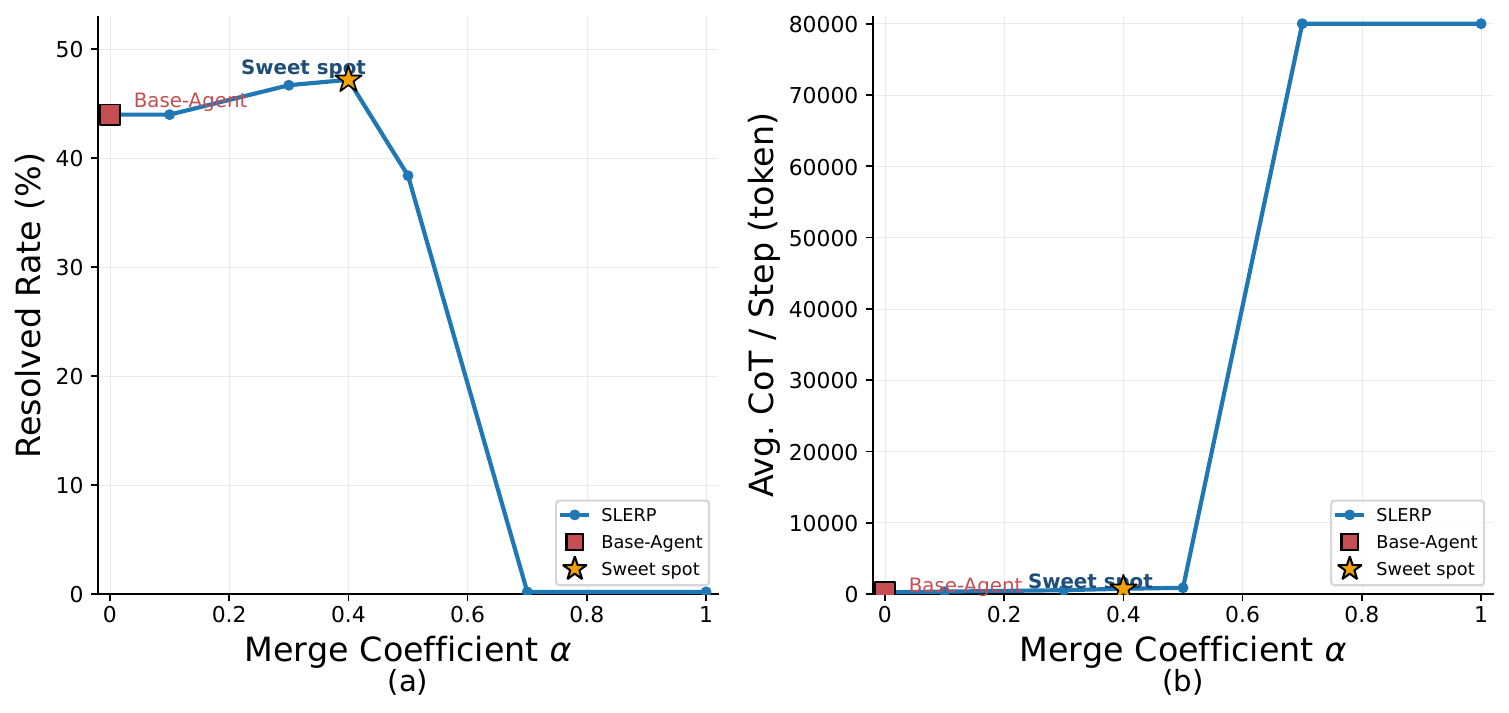}
    \caption{The effect of merge coefficient for SLERP. (a) SWE-Bench Verified performance, (b) Avg.CoT length per step}
    \label{fig:slerp_alpha_sweep}
\end{figure}

\begin{figure}[htb]
    \centering
    \includegraphics[width=1.0\linewidth]{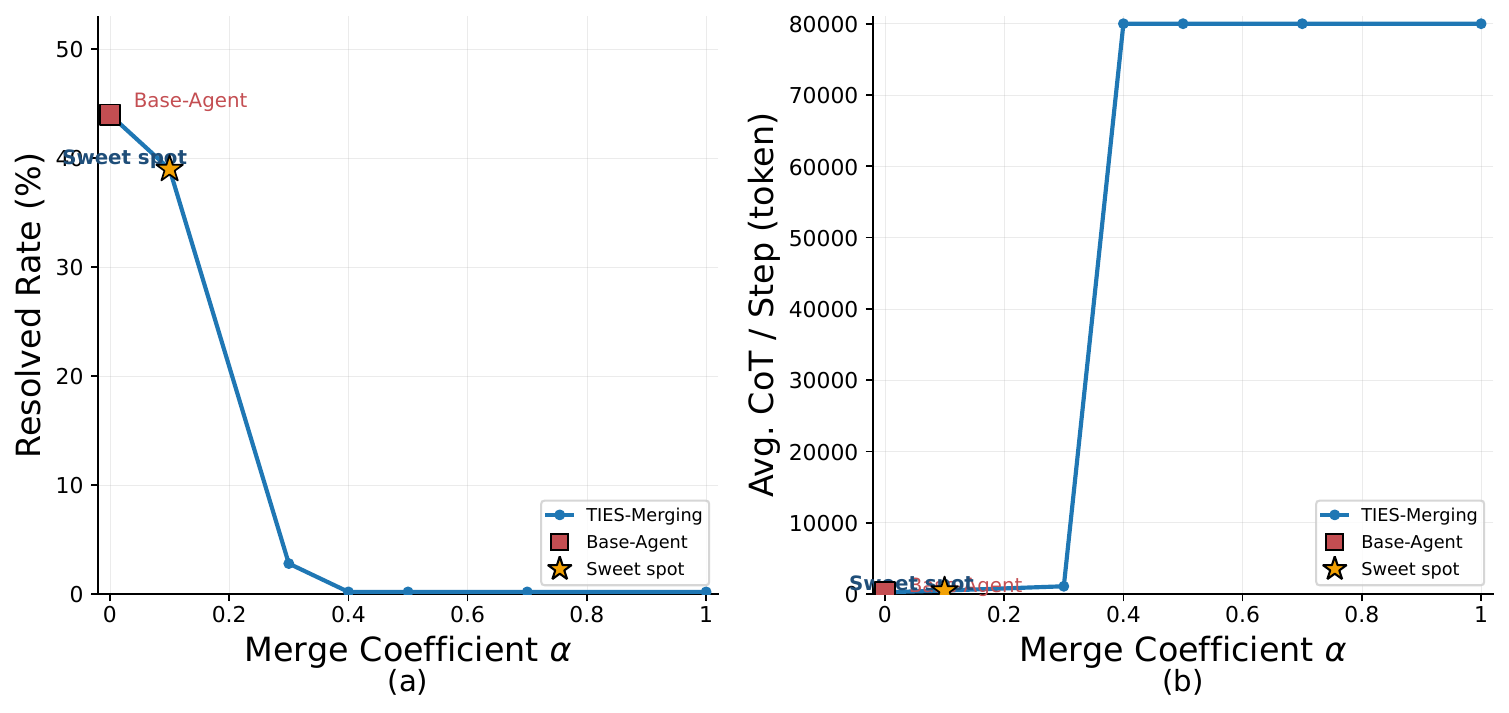}
    \caption{The effect of merge coefficient for TIES-Merging. (a) SWE-Bench Verified performance, (b) Avg.CoT length per step}
    \label{fig:ties_alpha_sweep}
\end{figure}

\begin{figure}[htb]
    \centering
    \includegraphics[width=1.0\linewidth]{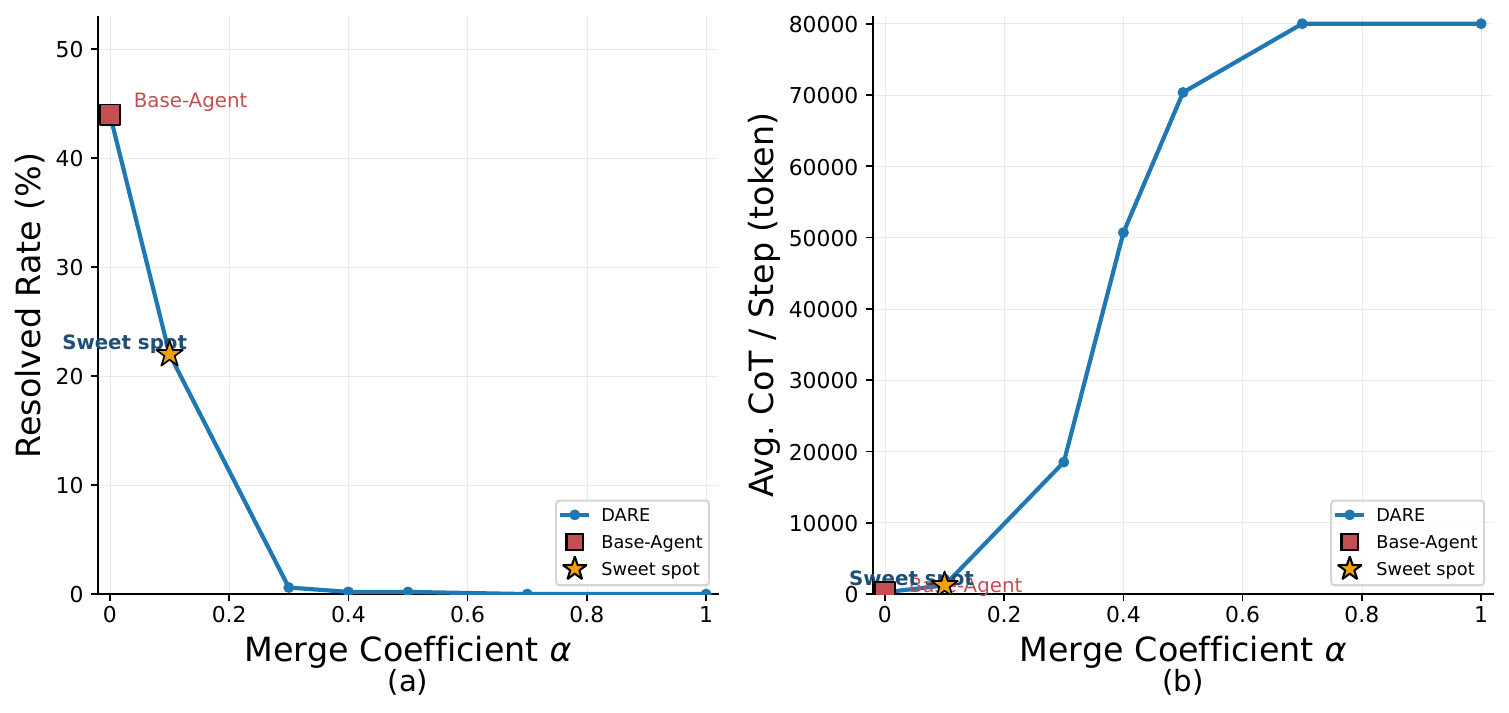}
    \caption{The effect of merge coefficient for DARE. (a) SWE-Bench Verified performance, (b) Avg.CoT length per step}
    \label{fig:dare_alpha_sweep}
\end{figure}



For all model-merging baselines, the source models are fixed to Agent-8B and Reasoning-8B. 
To ensure a fair comparison, we conduct a merge-coefficient sweep for each baseline method and report the best-performing sweet spot in the main results. 
Thus, the baseline numbers in Table~\ref{tab:main_results_full} are not obtained from arbitrary default coefficients, but from the strongest setting found under the same SWE-Bench Verified evaluation protocol.

Figures~\ref{fig:task_arithmetic_alpha_sweep}--\ref{fig:dare_alpha_sweep} show the merge-coefficient sweeps for Task Arithmetic, SLERP, TIES-Merging, and DARE. 
A consistent pattern is that these baselines are highly sensitive to the merge coefficient $\alpha$. 
When $\alpha$ is small, the injected reasoning signal is weak and the model remains close to the original Agent-8B. 
As $\alpha$ increases to a narrow sweet spot, Task Arithmetic and SLERP can obtain moderate improvements. 
However, further increasing $\alpha$ quickly destabilizes the agent: the model becomes dominated by the single-turn mathematical reasoning pattern, produces extremely long reasoning traces, and often reaches the maximum response-length limit. 
This causes the agent to stop interacting effectively with the environment, leading to a sharp collapse in SWE-Bench Verified performance.

This phenomenon highlights a key limitation of standard model merging. 
Although the merge coefficient $\alpha$ nominally controls the amount of reasoning-model update being added, it does not provide reliable control over the resulting behavior pattern. 
The average CoT length of the merged model does not scale smoothly or linearly with $\alpha$; instead, it often stays close to the base agent in the low-coefficient region and then abruptly jumps to near the response-length limit. 
Such discontinuous behavior makes baseline merging difficult to tune and easy to collapse, especially in agentic settings where the model must repeatedly alternate between thinking, acting, and observing.

In contrast, M2A exposes the global merge strength $\beta$ as an effective behavior-control knob. 
As shown in Figure~\ref{fig:beta_sweep}, increasing $\beta$ leads to an approximately monotonic, and nearly linear in the high-performing region, growth of Avg. CoT per step. 
This means that M2A can translate a simple parameter-space coefficient into predictable behavior-space control over reasoning intensity. 
Moreover, M2A has a stable medium-reasoning region where performance remains consistently strong, reducing the burden of manual hyperparameter tuning. 
The comparison between Figures~\ref{fig:beta_sweep} and \ref{fig:task_arithmetic_alpha_sweep}--\ref{fig:dare_alpha_sweep} therefore further demonstrates the advantage of M2A: it does not merely find a better merge coefficient, but changes the merge coefficient from a fragile tuning variable into a controllable interface for shaping post-merge agent behavior.

For RAIN-Merging, we follow the parameter setting of the original method. 
Specifically, RAIN-Merging does not use the same manually swept scalar coefficient as Task Arithmetic, SLERP, TIES-Merging, or DARE. 
Instead, its merge coefficients are \textbf{adaptively selected} by the instruction-attention guided merging coefficient mechanism proposed in the paper \cite{huang2026rain}. 
We keep this design unchanged to faithfully evaluate RAIN-Merging under its intended configuration.

\subsection{Average interaction steps under different merge strengths.}
\begin{figure}[htb]
    \centering
    \includegraphics[width=1.0\linewidth]{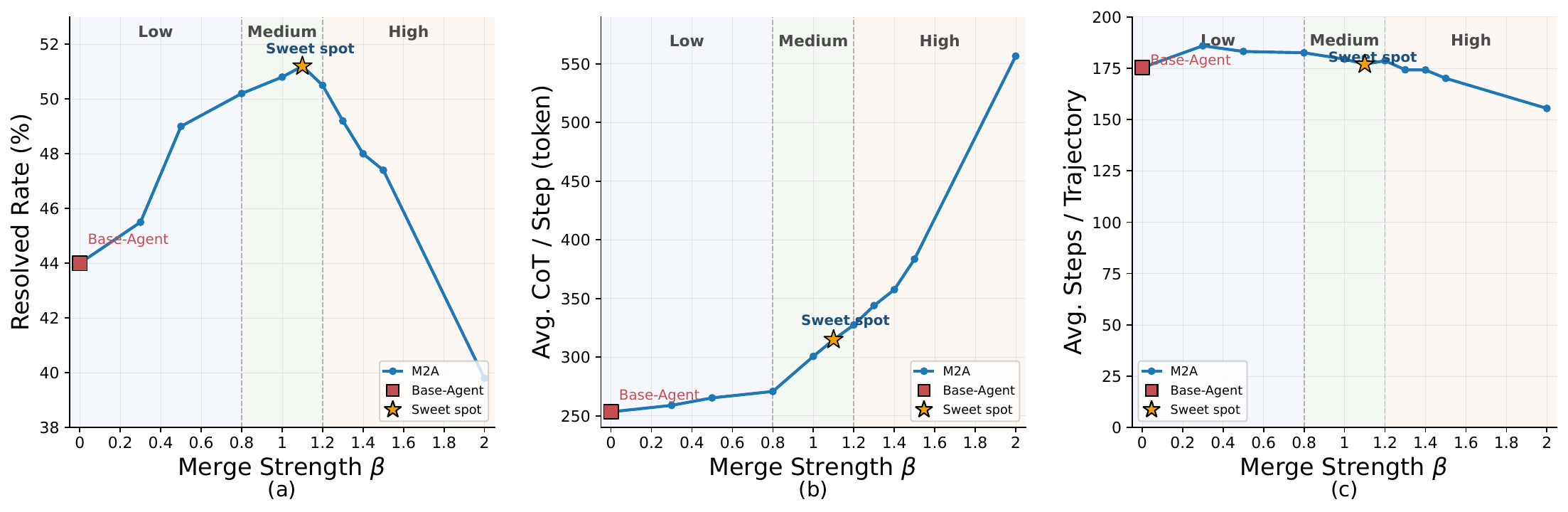}
    \caption{
    Effect of merge strength $\beta$ on merged model behavior. (a) SWE-Bench Verified performance, (b) average reasoning length per step, and (c) average interaction steps per trajectory.
    }
    \label{fig:app_beta_avg_step}
\end{figure}

We further analyze how merge strength $\beta$ affects the number of agent--environment interaction steps.
As shown in Figure~\ref{fig:app_beta_avg_step}, the average number of steps per trajectory remains \textbf{relatively stable} across the sweep, especially in the Low and Medium Reasoning regimes.
This indicates that increasing $\beta$ mainly controls the depth of internal reasoning, rather than suppressing the agent's external interaction behavior.
In the Medium Reasoning regime, where M2A achieves the best performance, the agent maintains a step count close to the Base Agent while producing longer and more useful reasoning traces.
Therefore, the performance gain does not come from reducing interaction or prematurely terminating the trajectory, but from better coordinating reasoning with action.

When $\beta$ enters the High Reasoning regime, the average step count begins to decrease mildly due to exceeding maximum prompt length.
Nevertheless, the step-count variation is much smoother than the abrupt behavior collapse observed in other merging baselines, further showing that M2A preserves the think--act--observe loop while exposing $\beta$ as a control knob for reasoning behavior.

\subsection{Additional Generalization Results}
\label{app:additional_generalization}

We provide comprehensive comparisons with additional
model-merging baselines, including SLERP, DARE, and
TIES-Merging, across different model families, agent
scaffolds, and benchmarks. The complete results are
reported in Table~\ref{tab:generalization_full}.
M2A consistently achieves the highest performance
across all evaluated settings, outperforming all
five model-merging baselines. Most conventional
merging methods substantially increase reasoning
length while reducing interaction steps, often
leading to degraded agent performance. In contrast,
M2A achieves stronger reasoning capabilities while
maintaining effective multi-turn interaction,
further demonstrating its robustness across
model families, agent scaffolds, and task domains.

\begin{table}[htbp]
\centering
\caption{Complete generalization results across model
families, agent scaffolds, and tasks. SWE denotes
SWE-Bench Verified. The best score in each setting
is highlighted in bold.}
\label{tab:generalization_full}

\begingroup
\footnotesize
\setlength{\tabcolsep}{5pt}
\renewcommand{\arraystretch}{0.95}

\begin{tabular*}{\linewidth}{
@{\extracolsep{\fill}}lccc@{}}
\toprule
\textbf{Method}
& \textbf{Score (\%) $\uparrow$}
& \textbf{Avg.\ Reasoning Length}
& \textbf{Avg. Step} \\
\midrule

\multicolumn{4}{@{}l}{
\textit{Llama-3.1-8B / OpenHands / SWE}} \\
Agent-8B
    & 17.6 & 207.79 & 67.66 \\
Task Arithmetic
    & 21.0 & 483.38 & 43.73 \\
TIES-Merging
    & 20.6 & 257.36 & 53.10 \\
DARE
    & 12.6 & 815.49 & 30.70 \\
SLERP
    & 21.8 & 250.68 & 61.70 \\
RAIN-Merging
    & 20.4 & 242.45 & 70.00 \\
\textbf{M2A}
    & \textbf{23.0} & 228.76 & 70.70 \\

\midrule
\addlinespace[3pt]
\multicolumn{4}{@{}l}{
\textit{Qwen3-8B / Mini-SWE-Agent / SWE}} \\
Agent-8B
    & 42.6 & 179.08 & 168.00 \\
Task Arithmetic
    & 43.5 & 866.42 & 80.79 \\
TIES-Merging
    & 38.6 & 420.12 & 92.25 \\
DARE
    & 23.8 & 1332.48 & 67.00 \\
SLERP
    & 33.8 & 707.38 & 83.25 \\
RAIN-Merging
    & 35.8 & 179.40 & 189.56 \\
\textbf{M2A}
    & \textbf{46.4} & 294.97 & 169.99 \\

\midrule
\addlinespace[3pt]
\multicolumn{4}{@{}l}{
\textit{Qwen3-8B / R2E / SWE}} \\
Agent-8B
    & 46.6 & 208.93 & 149.27 \\
Task Arithmetic
    & 48.7 & 847.47 & 70.70 \\
TIES-Merging
    & 47.0 & 513.47 & 87.10 \\
DARE
    & 34.6 & 1130.82 & 58.00 \\
SLERP
    & 46.8 & 715.84 & 76.00 \\
RAIN-Merging
    & 46.8 & 209.30 & 151.70 \\
\textbf{M2A}
    & \textbf{49.8} & 277.76 & 166.33 \\

\midrule
\addlinespace[3pt]
\multicolumn{4}{@{}l}{
\textit{Qwen3-8B / GAIA}} \\
Agent-8B
    & 73.33 & 850.00 & 13.65 \\
Task Arithmetic
    & 66.10 & 2759.10 & 11.04 \\
TIES-Merging
    & 66.67 & 2535.51 & 11.99 \\
DARE
    & 40.61 & 2571.79 & 9.51 \\
SLERP
    & 26.67 & 1891.57 & 11.58 \\
RAIN-Merging
    & 69.10 & 2007.10 & 14.45 \\
\textbf{M2A}
    & \textbf{78.79} & 959.00 & 19.22 \\

\bottomrule
\end{tabular*}
\endgroup
\end{table}

\section{Calibration Set Construction}
\label{app:calibration_set}

\subsection{Agent Calibration Set}
\label{app:agent_calibration_set}

We construct the calibration set from on-policy
agent trajectories on Scale-SWE~\cite{zhao2026immersion}.
We randomly sampled around 300 distinct tasks from Scale-SWE~\cite{zhao2026immersion} and used our SFT-trained Agent-8B model to solve them, resulting in 300 successful or failed task trajectories. Each trajectory consists of multiple rounds of think–action–observation interactions. We sampled 100 individual interaction steps from these 300 trajectories and used them as calibration data. Thus, the calibration samples cover different tasks, both successful and failed trajectories, and different interaction step within each trajectory, thereby reducing potential sampling bias. The calibration
set is used only to extract behavior-critical
hidden states for M2A, not for fine-tuning or
benchmark evaluation.

We verify that these 100 calibration trajectories have no instance-level overlap with SWE-Bench Verified. 
In particular, none of the calibration trajectories corresponds to any of the 500 SWE-Bench Verified evaluation instances. 
This separation ensures that the calibration stage does not expose the model to evaluation issues or solution trajectories from the test benchmark.

Each calibration example is organized into three fields: \texttt{prompt}, \texttt{reasoning}, and \texttt{response}. 
The \texttt{prompt} contains the system prompt and the multi-turn conversation history before the current action. 
The \texttt{reasoning} field contains the model's current-step internal thinking process, enclosed by \texttt{<think>} and \texttt{</think>}. 
The \texttt{response} field contains the model's current-step external action, usually a tool call. 
Figure~\ref{fig:calibration_format} gives a simplified example of the trajectory format.

\begin{figure}[h]
\centering
\small
\begin{minipage}{0.94\linewidth}
\begin{verbatim}
{
  "prompt": "<|im_start|>system
You are OpenHands agent, a helpful AI assistant that can interact 
with a computer to solve tasks.
...
<|im_start|>user
Now consider the following Github issue:
...
<|im_start|>assistant",

  "reasoning": "<think>
I need to understand the issue, inspect the repository, locate the
relevant implementation, and decide which file should be modified.
</think>",

  "response": "<tool_call>
{\"name\": \"execute_bash\", 
 \"arguments\": {\"command\": \"find /workspace/repo -type f | head\"}}
</tool_call>"
}
\end{verbatim}
\end{minipage}
\caption{\textbf{Simplified example of a code-agent calibration trajectory.} 
The prompt contains the system prompt and conversation history; the reasoning field contains the current-step internal thinking; and the response field contains the external tool call.}
\label{fig:calibration_format}
\end{figure}

\subsection{Behavior Marker Set}
\label{app:behavior_marker_set}

Each calibration example is first formatted with the model's chat template and then represented as
the concatenation of the prompt, reasoning field, and response field. We locate behavior-transition
markers in the tokenized sequence using character-to-token offset mapping. The marker set is not tied
to a specific serialization format; instead, it is instantiated according to the tool-call format used by
the agent trajectory. For XML-style tool calls, we use
\begin{equation}
\mathcal{M}_{\mathrm{xml}} =
\{\texttt{<think>}, \texttt{</think>}, \texttt{<function=}, \texttt{</function>}\},
\end{equation}
while for native tool-call formats, we use
\begin{equation}
\mathcal{M}_{\mathrm{native}} =
\{\texttt{<think>}, \texttt{</think>}, \texttt{<tool\_call>}, \texttt{</tool\_call>}\}.
\end{equation}
When mixed formats are present, M2A uses the union of all supported markers,
$\mathcal{M}=\mathcal{M}_{\mathrm{xml}}\cup\mathcal{M}_{\mathrm{native}}$.

For a start marker at token position $t$, we collect the local neighborhood $[t,t+r]$; for an end marker, we collect $[t-r,t]$. In the main experiments, we use $r=5$. This neighborhood construction captures the local representation changes when the model enters reasoning, exits reasoning, starts a tool call, or finishes an external action block.


\subsection{Calibration Set Size Ablation}
\label{app:calibration_size}

We further study the effect of calibration set size on M2A. 
All experiments use the same model, merge configuration, evaluation scaffold, and SWE-Bench Verified evaluation protocol; we only vary the number of calibration trajectories used to construct the agent-critical behavior subspace. 
As shown in Table~\ref{tab:calibration_size}, using only 50 trajectories already improves over the base agent, but the performance is lower than the default setting. 
Increasing the calibration set size to 100 trajectories yields the best resolved rate of 51.2\%. 
Further increasing the calibration set to 200 or 300 trajectories does not bring additional improvement, while requiring more calibration computation.

These results indicate that 100 trajectories are sufficient to capture the key behavior-critical directions for preserving the agent's think--act--observe pattern. 
Therefore, we use 100 calibration trajectories in the main experiments, which provides the best trade-off between performance and efficiency.

\begin{table}[htb]
  \centering
  \small
  \caption{Calibration set size ablation on SWE-Bench Verified.}
  \label{tab:calibration_size}
  \begin{tabular}{cc}
    \toprule
    \textbf{Calibration Set Size} & \textbf{Resolved Rate (\%)} \\
    \midrule
    50  & 49.1 \\
    100 & \textbf{51.2} \\
    200 & 51.0 \\
    300 & 50.8 \\
    \bottomrule
  \end{tabular}
\end{table}

\subsection{Calibration Window Radius Ablation}
\label{app:calibration_radius}

We further ablate the calibration window radius $r$, which determines how many neighboring tokens are collected around each behavior transition marker when constructing the agent-critical behavior subspace. 
This experiment examines whether M2A is sensitive to the local window used for behavior calibration. 
A very small window may miss the transition context around reasoning and tool-call markers, while an overly large window may include task-specific semantic tokens that are less directly related to the agent's think--act--observe behavior pattern.

All experiments use the same model, calibration set size, merge configuration, and SWE-Bench Verified evaluation protocol; we only vary the radius $r$. 
As shown in Table~\ref{tab:calibration_radius}, M2A is relatively robust to this hyperparameter. 
The resolved rate remains consistently high across different radius choices: $r=3$ achieves 51.0\%, $r=5$ achieves 51.2\%, and $r=10$ still obtains 49.5\%. 
All settings substantially outperform the base agent, indicating that the behavior-critical subspace can be reliably estimated from local marker neighborhoods and does not depend on a fragile radius choice.

At the same time, a moderate radius performs best. 
The slightly lower performance at $r=3$ suggests that a smaller window may not fully capture the local transition context, while the drop at $r=10$ suggests that larger windows may introduce more task-specific or irrelevant trajectory tokens into the protected subspace. 
Therefore, we use $r=5$ in the main experiments, as it provides the best performance while preserving robustness across nearby radius choices.

\begin{table}[htb]
  \centering
  \small
  \caption{\textbf{Calibration window radius ablation on SWE-Bench Verified.}}
  \label{tab:calibration_radius}
  \begin{tabular}{cc}
    \toprule
    \textbf{Window Radius $r$} & \textbf{Resolved Rate (\%)} \\
    \midrule
    3  & 51.0 \\
    5  & \textbf{51.2} \\
    10 & 49.5 \\
    \bottomrule
  \end{tabular}
\end{table}

\section{Compute Resources}
\label{app:compute_resources}

Table~\ref{tab:compute_resources} reports the compute resources used in our experiments. 
M2A itself is training-free and does not require gradient updates, optimizer states, or rollout-based policy optimization. 
Its additional cost comes from forward-pass calibration and parameter-space merging.

\begin{table}[h]
\centering
\small
\caption{\textbf{Compute resources.} We report the hardware, peak memory usage, and Wall-clock Time used for SFT and M2A merging.}
\label{tab:compute_resources}
\begin{tabular}{lccc}
\toprule
\textbf{Stage} & \textbf{Hardware} & \textbf{Peak Memory / GPU} & \textbf{Time(hours)} \\
\midrule
Agent-8B SFT 
& 32 $\times$ H200 
& 140 GB 
& $\approx 30$ \\

Multi-Task-8B SFT 
& 32 $\times$ H200 
& 140 GB 
& $\approx 48$ \\

M2A calibration and merging 
& 8 $\times$ H200 
& 50 GB 
& $\approx 2$ \\

\bottomrule
\end{tabular}
\end{table}

\section{Implementation Details of M2A}
\label{app:m2a_implementation}

\paragraph{Orthonormal basis for the agent-critical subspace.}
The main text defines the agent-critical behavior subspace as
\begin{equation}
    S_{\mathrm{agent}}^{(l)} = \mathrm{span}(C^{(l)}),
    \qquad
    C^{(l)} = [h_1^{(l)}, h_2^{(l)}, \ldots, h_{n_l}^{(l)}] \in \mathbb{R}^{d \times n_l},
\end{equation}
where each $h_i^{(l)}$ is a hidden feature collected from the local neighborhood of a behavior-transition marker.
Conceptually, an orthonormal basis $Q^{(l)}$ of this subspace can be obtained by a thin SVD:
\begin{equation}
    C^{(l)} = U^{(l)} \Sigma^{(l)} V^{(l)\top}.
\end{equation}
We keep the left singular vectors corresponding to non-negligible singular values:
\begin{equation}
    Q^{(l)}
    =
    U^{(l)},
\end{equation}
In this explicit form, the null-space projector is
\begin{equation}
    P_{\mathrm{null}}^{(l)}
    =
    I - Q^{(l)}Q^{(l)\top},
\end{equation}
and the behavior-preserving reasoning update for a weight matrix whose input dimension matches $S_{\mathrm{agent}}^{(l)}$ is:
\begin{equation}
    \widetilde{\Delta W}_{\mathrm{reason}}^{(l)}
    =
    \Delta W_{\mathrm{reason}}^{(l)}
    P_{\mathrm{null}}^{(l)}.
\end{equation}
This yields
\begin{equation}
    \widetilde{\Delta W}_{\mathrm{reason}}^{(l)} h = 0,
    \qquad
    \forall h \in S_{\mathrm{agent}}^{(l)},
\end{equation}
which is the no-perturbation condition used in the main text.

\paragraph{Matrix-free implementation used in experiments.}
Although the above SVD form is useful for presenting the geometry of M2A, our implementation does not explicitly materialize $Q^{(l)}$ or $P_{\mathrm{null}}^{(l)}$.
This is mainly for efficiency: in a transformer layer, the protected constraints must be applied separately to multiple parameter blocks, including $W_Q$, $W_K$, $W_V$, $W_O$, $W_{\mathrm{gate}}$, $W_{\mathrm{up}}$, and $W_{\mathrm{down}}$, and for attention projections the update is further processed head-wise. In the default setting, we merge all these components, denoted as \texttt{qkvof}. 


Instead, we implement the same null-space projection through block-wise linear constraint operators.
For a mergeable block $B$ in layer $l$, let
\begin{equation}
    \delta_B = \mathrm{vec}(\Delta W_B)
\end{equation}
be the vectorized reasoning task update.
We construct a linear operator $A_B$ from the same marker-neighborhood calibration features.
The protected-subspace condition is then written as:
\begin{equation}
    A_B \widetilde{\delta}_B = 0.
\end{equation}
The projected update is obtained by solving:
\begin{equation}
    \widetilde{\delta}_B
    =
    \arg\min_z
    \|z-\delta_B\|_2^2
    \quad
    \mathrm{s.t.}
    \quad
    A_B z = 0.
\end{equation}
The closed-form solution is:
\begin{equation}
    \widetilde{\delta}_B
    =
    \delta_B
    -
    A_B^\top
    (A_B A_B^\top)^{-1}
    A_B \delta_B .
\end{equation}
For numerical stability, we use the ridge-stabilized form:
\begin{equation}
    \widetilde{\delta}_B
    =
    \delta_B
    -
    A_B^\top
    (A_B A_B^\top + \lambda I)^{-1}
    A_B \delta_B ,
\end{equation}
where $\lambda=10^{-4}$ by default.

This matrix-free formulation is equivalent to the explicit $Q$-based projection when $A_B$ spans the same protected feature subspace.
In practice, $A_B$ is built from sampled marker-neighborhood constraints, so it provides an efficient sampled approximation to the ideal projection
$\Delta W_{\mathrm{reason}}^{(l)}(I-Q^{(l)}Q^{(l)\top})$.
The implementation only requires products with $A_B$ and $A_B^\top$, avoiding the memory cost of constructing the full projector for every layer, head, and parameter block.

\paragraph{Block-wise projection details.}
The implementation applies the above projection block-wise.
For attention projections, $W_Q$, $W_K$, and $W_V$ are sliced by head rows, while $W_O$ is sliced by head columns.
For models with grouped-query attention, the key/value head corresponding to query head $h$ is selected by $h \bmod H_{\mathrm{kv}}$.
For FFN layers, the same projection principle is applied to $W_{\mathrm{gate}}$, $W_{\mathrm{up}}$, and $W_{\mathrm{down}}$; the down-projection update is internally transposed to match the solver layout and then transposed back before being added to the model.

Small constraint systems are solved by dense linear solvers.
When dense solving is unstable or memory-expensive, we fall back to conjugate gradient on
$(A_BA_B^\top+\lambda I)$ with a maximum of 100 iterations and tolerance $10^{-5}$.
Thus, the implementation follows the null-space projected merging objective in the main text, but realizes it through an efficient matrix-free solver rather than explicitly computing $Q^{(l)}$ for every block.

\subsection{Implementation Details of Adaptive Layer-Wise Merging}
\label{app:adaptive_layerwise_details}

\paragraph{Layer-wise task-vector aggregation.}
In the main text, $\Delta\theta_{\mathrm{agent}}^{(l)}$ and
$\Delta\theta_{\mathrm{reason}}^{(l)}$ denote the layer-wise agent and reasoning
task vectors. In implementation, each of them is represented as a collection of
mergeable parameter blocks:
\begin{equation}
    \Delta\theta_x^{(l)}
    =
    \left\{
    \Delta W_{x,p}^{(l)}
    \right\}_{p\in\mathcal{P}^{(l)}},
    \qquad
    x\in\{\mathrm{agent},\mathrm{reason}\},
\end{equation}
where
\begin{equation}
    \Delta W_{\mathrm{agent},p}^{(l)}
    =
    W_{\mathrm{agent},p}^{(l)}
    -
    W_{\mathrm{base},p}^{(l)},
    \qquad
    \Delta W_{\mathrm{reason},p}^{(l)}
    =
    W_{\mathrm{reason},p}^{(l)}
    -
    W_{\mathrm{base},p}^{(l)}.
\end{equation}
Here $\mathcal{P}^{(l)}$ contains all enabled mergeable blocks in layer $l$,
including the Q/K/V/O attention projections and the FFN gate/up/down projections.
For attention modules, Q/K/V/O are computed head-wise; for FFN modules, the full
projection matrices are used.

\begin{figure}[htb]
    \centering
    \includegraphics[
        width=1\linewidth,
        trim=150 400 200 0,
        clip
    ]{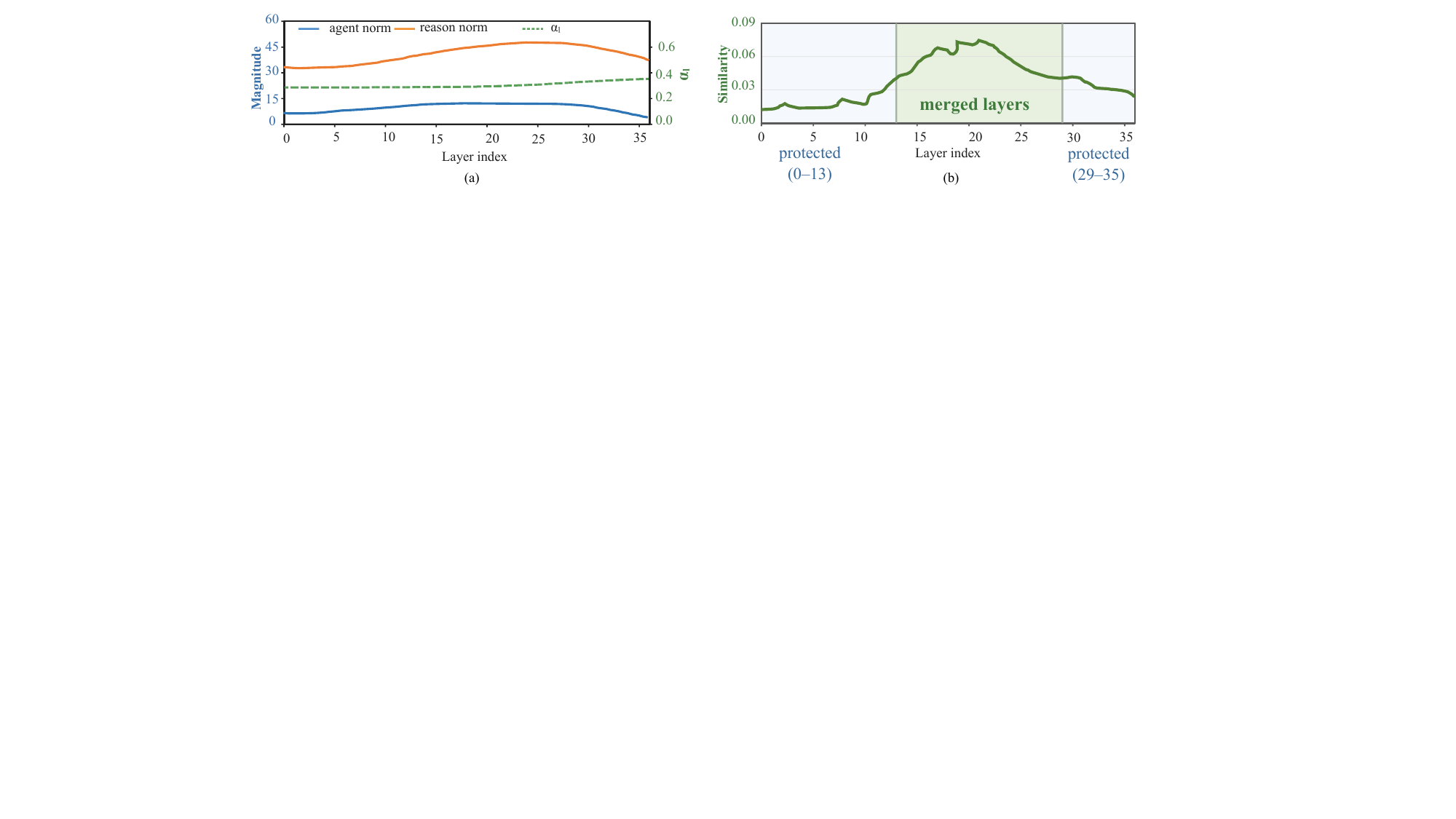}
    \caption{(a) $\alpha_l$ is computed from the ratio between the layer-wise agent and
    reasoning task-vector norms.
    (b) $M_l$ is obtained by layer-wise similarities. Green regions are merged layers, while blue regions are protected layers.
    }
    \label{fig:app_alpha_mask}
\end{figure}

\paragraph{Merge coefficient calibration.}
The layer-wise Frobenius norm in Eq.~\eqref{eq:alpha} is computed by summing the
squared norms over all enabled blocks:
\begin{equation}
    \left\|
    \Delta\theta_x^{(l)}
    \right\|_F
    =
    \left(
    \sum_{p\in\mathcal{P}^{(l)}}
    \left\|
    \Delta W_{x,p}^{(l)}
    \right\|_F^2
    \right)^{1/2},
    \qquad
    x\in\{\mathrm{agent},\mathrm{reason}\}.
    \label{eq:app_layer_norm}
\end{equation}
The calibrated coefficient is then computed as
\begin{equation}
    \alpha_l
    =
    \beta
    \cdot
    \frac{
    \left\|
    \Delta\theta_{\mathrm{agent}}^{(l)}
    \right\|_F
    }{
    \left\|
    \Delta\theta_{\mathrm{reason}}^{(l)}
    \right\|_F + \epsilon
    }.
    \label{eq:app_alpha_impl}
\end{equation}
If the reasoning-vector norm is numerically close to zero, we set $\alpha_l=0$.
This is the quantity visualized in Figure~\ref{fig:app_alpha_mask}(a): the
reasoning task vector often has a larger magnitude than the agent task vector,
and $\alpha_l$ rescales it to the layer-wise agent scale before merging.

\paragraph{Similarity-aware layer mask.}
In the main text, $\mathrm{sim}^{(l)}$ denotes the cosine similarity between
$\Delta\theta_{\mathrm{agent}}^{(l)}$ and
$\Delta\theta_{\mathrm{reason}}^{(l)}$. In implementation, this similarity is
computed by aggregating all enabled mergeable blocks in layer $l$:
\begin{equation}
    \mathrm{sim}^{(l)}
    =
    \frac{
    \sum_{p\in\mathcal{P}^{(l)}}
    \left\langle
    \Delta W_{\mathrm{agent},p}^{(l)},
    \Delta W_{\mathrm{reason},p}^{(l)}
    \right\rangle_F
    }{
    \left\|
    \Delta\theta_{\mathrm{agent}}^{(l)}
    \right\|_F
    \left\|
    \Delta\theta_{\mathrm{reason}}^{(l)}
    \right\|_F
    + \epsilon
    },
    \label{eq:app_sim_impl}
\end{equation}
where $\mathcal{P}^{(l)}$ denotes the set of mergeable parameter blocks in layer
$l$, and $\langle A,B\rangle_F=\mathrm{Tr}(A^\top B)$ is the Frobenius inner
product. The global similarity threshold is the mean similarity over all selected
layers:
\begin{equation}
    \bar{\mathrm{sim}}
    =
    \frac{1}{|\mathcal{L}|}
    \sum_{l\in\mathcal{L}}
    \mathrm{sim}^{(l)},
    \label{eq:app_mean_sim}
\end{equation}
where $\mathcal{L}$ is the set of selected layers. The binary layer mask is computed as
\begin{equation}
    M_l
    =
    \mathbb{I}
    \left[
    \mathrm{sim}^{(l)}
    >
    \bar{\mathrm{sim}}
    \right].
    \label{eq:app_mask_impl}
\end{equation}
Layers with $M_l=1$ are merged after null-space projection, while layers with
$M_l=0$ are skipped and kept identical to the agent model. As illustrated in
Figure~\ref{fig:app_alpha_mask}(b), this mask selects similarity-aligned layers
for reasoning injection and protects low-similarity layers from potentially
conflicting updates.

\subsection{Calibration Source Quality and Scaffold Distribution}
\label{app:calibration_sources}
\begin{table}[htbp]
\centering\small
\caption{Calibration-source comparison on Qwen3-8B/OpenHands with SWE-Bench Verified. Mixed-scaffold calibration uses 50\% Mini-SWE-Agent and 50\% R2E trajectories; the resulting merged model is evaluated with OpenHands.}
\begin{tabularx}{\textwidth}{@{}Xrrr@{}}
\toprule
Calibration source & Score (\%) & Avg.Reasoning Length & Avg.Step \\
\midrule
Random sampling & \textbf{51.2} & 327.40 & 178.00 \\
Failed rollouts & 50.6 & 299.29 & 165.77 \\
Successful rollouts & 50.2 & 291.53 & 162.59 \\
Mixed scaffolds & 50.0 & 302.90 & 166.67 \\
\bottomrule
\end{tabularx}
\end{table}
For the mixed-scaffold setting, we constructed the calibration set using trajectories from Mini-SWE-Agent and R2E, performed merging with this calibration set, and evaluated the resulting model with OpenHands. Across all calibration sources, the performance differences are consistently small, indicating that M2A is not sensitive to the specific choice of trajectories, whether they are successful, failed, or from different scaffolds. Random sampling performs slightly better, because it provides greater trajectory diversity. Overall, these results support the robustness of M2A to both calibration-source quality and scaffold distribution.

\subsection{Alternative Layer-Selection Rules}
\label{app:layer_rule}
\begin{table}[htbp]
\centering\small
\caption{Layer-selection ablation on Qwen3/OpenHands. Layer ranges use the reported indices and are not counts. The global-mean threshold is approximately 0.03.}
\begin{tabularx}{\textwidth}{@{}Xlrrr@{}}
\toprule
Rule & Selected layers & Score (\%) & Avg.Reasoning Length & Avg.Step \\
\midrule
Global mean ($\approx0.03$) & 14--28 & \textbf{51.2} & 327.40 & 178.00 \\
Top-5 & 18--22 & 46.2 & 247.49 & 159.49 \\
Top-10 & 16--25 & 45.6 & 292.54 & 151.71 \\
Fixed threshold (0.04) & 15--25 & 49.6 & 283.15 & 165.34 \\
Fixed threshold (0.05) & 17--23 & 50.0 & 308.22 & 154.29 \\
\bottomrule
\end{tabularx}
\end{table}
The global-mean similarity rule performs better than the tested top-(k) alternatives and does not require an additional hyperparameter.
The global-mean rule outperforms Top-(5) and Top-(10) by 5.0 and 5.6 percentage points, respectively. It adapts automatically to the layer-similarity distribution and avoids introducing an additional (k) or manually tuned threshold.

\paragraph{Reproducibility and numerical details.}
All models are loaded in bfloat16, while projection computations are performed in float32 by default. The implementation supports assigning Q/K, V/O, and FFN projection computations to different devices. To improve robustness for long trajectories, calibration examples exceeding the maximum sequence length are skipped during constraint construction.

\paragraph{Chat template for multi-turn reasoning.}
Qwen3 models use a default chat template that compresses historical assistant reasoning to save context length. 
Specifically, for previous assistant turns before the latest user query, the default template removes the content inside \texttt{<think>}...\texttt{</think>} and only keeps the final assistant response. 
While this design is efficient for ordinary dialogue, it can be suboptimal for multi-turn agent evaluation, because the agent may need to condition on its previous reasoning traces when planning subsequent tool calls, interpreting observations, or revising earlier decisions.

To better evaluate the effect of reasoning integration in multi-turn agent settings, we modify the chat template to preserve historical reasoning traces. 
The key change is that, for every assistant message, if \texttt{reasoning\_content} is available or can be parsed from \texttt{<think>}...\texttt{</think>}, we explicitly serialize it back into the conversation:
\begin{equation}
\texttt{assistant message}
\; \Rightarrow \;
\texttt{<think>}~\texttt{reasoning\_content}~\texttt{</think>}
\;+\;
\texttt{final content}.
\end{equation}
In other words, we remove the default logic that only preserves reasoning after the latest non-tool user query. 
The modified template therefore keeps the full multi-turn reasoning history, including previous thought traces, tool calls, and tool responses.

The tool-call format itself is unchanged. 
Function signatures are still provided inside \texttt{<tools>}...\texttt{</tools>}, tool invocations are serialized with \texttt{<tool\_call>}...\texttt{</tool\_call>}, and environment observations are returned inside \texttt{<tool\_response>}...\texttt{</tool\_response>}. 
Thus, the modification only affects whether historical reasoning traces are retained, not the tool-use protocol.

For fairness, all evaluated models use the same reasoning-preserving chat template. 
Therefore, the reported gains do not come from giving M2A a different conversation format. 
All models are evaluated under the same context-length limit, tool-call format, and trajectory construction protocol.


\section{Limitations}
\label{app:limitations}

Our main controlled study focuses on Qwen3-8B
with OpenHands and SWE-Bench Verified; additional
experiments cover Llama-3.1-8B, Mini-SWE-Agent,
R2E, and GAIA.
These settings support transfer beyond the original
configuration but do not establish universal
preservation across arbitrary tool schemas or
model scales.
In particular, interaction counts may increase
on longer-horizon tasks, and
calibration still requires observable action/turn
boundaries and access to model weights.
Broader deep-research settings and sensitivity
to implicit boundaries remain open directions.

\section{Broader Impact}
\label{app:impact}
This work studies a training-free method for synergizing mathematical and agentic reasoning in large language models.
Its potential positive impact is to improve the capability and efficiency of software engineering agents without requiring additional costly SFT or RL, making stronger agentic reasoning more accessible to researchers and practitioners with limited computational resources. 
By preserving multi-turn interaction behavior while enhancing internal reasoning, M2A may help build coding agents that perform more reliable debugging, code repair, and repository-level understanding.
At the same time, stronger coding agents may also introduce risks if deployed without proper safeguards. 
More capable software agents could be misused to automate the generation of insecure code, discover or exploit vulnerabilities, or make unintended modifications in real-world codebases. 
In addition, because agentic systems interact with external environments, failures may propagate beyond text generation and affect files, repositories, or execution environments. 
Therefore, practical deployment should include sandboxed execution, human oversight for consequential changes, access control, logging, and security-oriented evaluation before use in real-world software systems.


\clearpage
\section{Qualitative Trajectory Case Study}
\label{app:trajectory_case_study}
\subsection{Case Study 1}
To better illustrate the behavioral change induced by M2A, we present a concrete SWE-Bench Verified case study from the same issue instance, where Agent-8B fails while M2A-Agent-8B succeeds. 
Both agents are evaluated under the same OpenHands scaffold and are given the same repository-level issue. 
The issue requires fixing a Sphinx autodoc bug: when Google-style docstrings are parsed by Napoleon, the return description is represented as \texttt{:returns:}, but the autodoc type-hint merger only recognizes \texttt{:return:} when deciding whether to add \texttt{:rtype:} under \texttt{autodoc\_typehints\_description\_target = "documented"}.

\begin{center}
\begin{tcolorbox}[
  enhanced,
  breakable,
  width=\textwidth,
  colback=white,
  colframe=m2aDarkGray!45,
  boxrule=0.6pt,
  arc=2pt,
  title=\textbf{Trajectory comparison on \texttt{sphinx-doc/sphinx}: \texttt{autodoc\_typehints\_description\_target} with Napoleon},
  coltitle=white,
  colbacktitle=m2aDarkGray,
  fonttitle=\bfseries
]

\begin{tcolorbox}[
  enhanced,
  breakable,
  colback=m2aRed!3,
  colframe=m2aRed!55,
  title=\textbf{Agent-8B baseline: failed trajectory},
  coltitle=white,
  colbacktitle=m2aRed!75!black,
  fonttitle=\bfseries,
  arc=2pt,
  boxrule=0.5pt
]
\trajtag{m2aRed}{Outcome: max prompt length exceeded}

\vspace{1mm}
\textbf{Pattern 1: broad exploration without a stable problem representation.}
\begin{lstlisting}[style=trajstyle]
<think>
The issue is about autodoc_typehints_description_target
not working with Napoleon. Maybe the bug is in how
Napoleon processes docstrings, or how autodoc merges fields.
Let me inspect more files and tests.
</think>

<tool_call>
grep -n "documented" /testbed/tests/test_ext_autodoc.py
grep -rn "Returns:" /testbed/tests/roots/test-ext-autodoc/
grep -rn "def _process_docstring" /testbed/sphinx/ext/
</tool_call>
\end{lstlisting}

\textbf{Pattern 2: repeated uncertainty and delayed action.}
\begin{lstlisting}[style=trajstyle]
<think>
Maybe there is no test covering Napoleon.
Maybe the issue is that :rtype: is added but removed later.
Maybe the output differs because of configuration.
I am going in circles.
</think>
\end{lstlisting}

\textbf{Pattern 3: context consumed before a reliable fix.}
\begin{lstlisting}[style=trajstyle]
<tool_call>
create /testbed/reproduce_issue.py
run reproduction
inspect Napoleon processing
inspect autodoc typehints
search tests again
...
</tool_call>

termination_reason = max_prompt_length_exceeded
reward = 0
\end{lstlisting}

\textbf{Interpretation.}
The baseline performs many exploratory actions but does not quickly compress the evidence into a precise root cause. 
Its trajectory is dominated by repeated search and hypothesis revision, causing the context to grow until the task fails.
\end{tcolorbox}

\begin{tcolorbox}[
  enhanced,
  breakable,
  colback=m2aGreen!3,
  colframe=m2aGreen!55,
  title=\textbf{M2A-Agent-8B: successful trajectory},
  coltitle=white,
  colbacktitle=m2aGreen!70!black,
  fonttitle=\bfseries,
  arc=2pt,
  boxrule=0.5pt
]
\trajtag{m2aGreen}{Outcome: resolved}

\vspace{1mm}
\textbf{Pattern 1: evidence-grounded problem representation.}
\begin{lstlisting}[style=trajstyle]
<think>
The bug is in documented-mode type-hint merging.
Napoleon converts Google-style "Returns:" into a
:returns: field, while autodoc checks only :return:
when deciding whether the return value is documented.
</think>

<tool_call>
grep -n "augment_descriptions_with_types" \
  /testbed/sphinx/ext/autodoc/typehints.py
</tool_call>
\end{lstlisting}

\textbf{Pattern 2: fewer but more targeted external actions.}
\begin{lstlisting}[style=trajstyle]
<edit>
--- a/sphinx/ext/autodoc/typehints.py
+++ b/sphinx/ext/autodoc/typehints.py

- elif parts[0] == 'return':
+ elif parts[0] in ('return', 'returns'):
      has_description.add('return')

- if name == 'return':
+ if name in ('return', 'returns'):
      continue
</edit>
\end{lstlisting}

\textbf{Pattern 3: verification before finalization.}
\begin{lstlisting}[style=trajstyle]
<tool_call>
python reproduce_issue.py
python -m pytest tests/test_ext_autodoc_configs.py \
  -k "description" -xvs
</tool_call>

Return type field FOUND.
1 passed.
termination_reason = env_done
reward = 1
\end{lstlisting}

\textbf{Interpretation.}
M2A first builds a compact causal explanation of the failure, then converts it into a minimal code change and verifies the result. 
This trajectory shows stronger internal reasoning without suppressing the external think--act--observe loop.
\end{tcolorbox}

\begin{tcolorbox}[
  enhanced,
  breakable,
  colback=m2aBlue!4,
  colframe=m2aBlue!45,
  boxrule=0.4pt,
  arc=2pt,
  title=\textbf{Key behavioral difference},
  coltitle=white,
  colbacktitle=m2aBlue!70!black,
  fonttitle=\bfseries
]
\small
The baseline follows an \trajtag{m2aRed}{explore--revise--explore} pattern: it repeatedly searches related code paths and tests, but fails to stabilize the root cause before exhausting the prompt budget. 
In contrast, M2A follows an \trajtag{m2aGreen}{understand--patch--verify} pattern: it identifies the semantic mismatch between \texttt{:return:} and \texttt{:returns:}, performs a localized patch, and validates the fix. 
This supports the trajectory-level finding that M2A shifts the agent from early trial-and-error editing toward evidence-grounded action.
\end{tcolorbox}

\end{tcolorbox}

\captionof{figure}{Qualitative trajectory comparison.
Agent-8B fails by repeatedly exploring related code paths without forming a stable root-cause representation. 
M2A-Agent-8B succeeds by first identifying the Napoleon--autodoc field mismatch, then applying a minimal patch and verifying it. 
The example illustrates how M2A strengthens internal reasoning while preserving effective external action.}
\label{fig:trajectory_case_study}
\end{center}

\subsection{Case Study 2}
\label{app:trajectory_case_sklearn}

We provide another representative trajectory comparison on a \texttt{scikit-learn} issue.
The issue concerns an inconsistency in \texttt{LogisticRegressionCV}:
when probabilistic scorers such as \texttt{neg\_log\_loss} are used with
\texttt{multi\_class='multinomial'}, the helper estimator inside
\texttt{\_log\_reg\_scoring\_path} should use the same multi-class mode as the
outer \texttt{LogisticRegressionCV}. This case illustrates the behavioral
difference between the base agent and M2A. The base agent identifies the
surface-level argument to add, but makes an incomplete edit that removes an
existing constructor argument. In contrast, M2A first builds a more precise
causal understanding of how scoring depends on \texttt{predict\_proba}, and then
translates this understanding into a minimal and semantics-preserving edit.

\begin{center}
\small

\begin{tcolorbox}[
    enhanced,
    breakable,
    width=\textwidth,
    colback=red!2,
    colframe=red!45!black,
    coltitle=white,
    colbacktitle=red!55!black,
    title=\textbf{Agent-8B Baseline: Trial-and-Error Edit},
    fonttitle=\bfseries,
    arc=2mm,
    boxrule=0.6pt,
    left=1.5mm,
    right=1.5mm,
    top=1mm,
    bottom=1mm
]
\textbf{Trajectory pattern.}
The base agent quickly identifies that \texttt{multi\_class} is relevant, but
does not fully preserve the semantics of the existing constructor call.
Its external action is close to the suggested fix, yet it overwrites the
existing \texttt{fit\_intercept} argument.

\vspace{2mm}
\textbf{Reasoning focus.}
\begin{itemize}
    \item Locates \texttt{\_log\_reg\_scoring\_path}.
    \item Recognizes that \texttt{multi\_class} should be passed.
    \item Applies the edit without preserving all existing arguments.
\end{itemize}

\vspace{1mm}
\textbf{Final patch.}
\begin{lstlisting}[style=trajcode]
- log_reg = LogisticRegression(fit_intercept=fit_intercept)
+ log_reg = LogisticRegression(multi_class=multi_class)
\end{lstlisting}

\vspace{1mm}
\textbf{Key failure pattern.}
The agent performs a near-correct but incomplete edit: it adds the new
argument by replacing the constructor call rather than extending it. As a
result, the patch loses \texttt{fit\_intercept=fit\_intercept}, which changes
the behavior of the helper estimator beyond the intended fix.

\vspace{1mm}
\textbf{Outcome.} Failed resolution.
\end{tcolorbox}

\vspace{2mm}

\begin{tcolorbox}[
    enhanced,
    breakable,
    width=\textwidth,
    colback=green!2,
    colframe=green!45!black,
    coltitle=white,
    colbacktitle=green!45!black,
    title=\textbf{M2A-Agent-8B: Evidence-Grounded Action},
    fonttitle=\bfseries,
    arc=2mm,
    boxrule=0.6pt,
    left=1.5mm,
    right=1.5mm,
    top=1mm,
    bottom=1mm
]
\textbf{Trajectory pattern.}
M2A first reasons about the scoring path and the probability prediction
mechanism before editing. It identifies that probabilistic scorers call
\texttt{predict\_proba}, whose behavior depends on the estimator's
\texttt{multi\_class} setting. This leads to a focused and semantics-preserving
patch.

\vspace{2mm}
\textbf{Reasoning focus.}
\begin{itemize}
    \item Traces \texttt{scoring(log\_reg, X\_test, y\_test)}.
    \item Connects \texttt{neg\_log\_loss} to \texttt{predict\_proba}.
    \item Infers that \texttt{multi\_class} must be inherited by the helper estimator.
    \item Preserves the existing \texttt{fit\_intercept} behavior.
\end{itemize}

\vspace{1mm}
\textbf{Final patch.}
\begin{lstlisting}[style=trajcode]
- log_reg = LogisticRegression(fit_intercept=fit_intercept)
+ log_reg = LogisticRegression(fit_intercept=fit_intercept,
+                              multi_class=multi_class)
\end{lstlisting}

\vspace{1mm}
\textbf{Key success pattern.}
The edit is not merely longer or more verbose reasoning. The additional
reasoning is used to construct a better problem representation, which is then
converted into a smaller and more faithful external action.

\vspace{1mm}
\textbf{Outcome.} Successful resolution.
\end{tcolorbox}

\vspace{2mm}

\captionof{figure}{
\textbf{Trajectory comparison on a \texttt{scikit-learn} issue.}
The base agent exhibits a trial-and-error pattern: it identifies the relevant
parameter but applies an incomplete edit that removes the existing
\texttt{fit\_intercept} argument. M2A instead performs evidence-grounded action:
it reasons through the scoring and probability-prediction path, then applies a
minimal patch that preserves the original constructor semantics while adding
\texttt{multi\_class}. This case supports our observation that M2A shifts agent
behavior from early, surface-level editing toward evidence-grounded action.
}
\label{fig:trajectory_case_sklearn}

\end{center}

\end{document}